\documentclass{article} 
\usepackage[table]{xcolor}
\usepackage{iclr2027_conference,times}
\usepackage{booktabs}
\usepackage{multirow}
\usepackage{subcaption}
\usepackage{graphicx}

\usepackage{amsmath,amsfonts,bm}

\def\eqref#1{equation~\ref{#1}}

\def\1{\bm{1}}

\DeclareMathAlphabet{\mathsfit}{\encodingdefault}{\sfdefault}{m}{sl}
\SetMathAlphabet{\mathsfit}{bold}{\encodingdefault}{\sfdefault}{bx}{n}

\usepackage{hyperref}
\usepackage{url}
\usepackage{booktabs,multirow}

\usepackage{wrapfig}
\definecolor{oursbg}{RGB}{215,235,230}
\title{Complementary Future Token Pruning for World Action Models}

\author{
Baekseung Kim$^{1}$ \hspace{2mm}
Junhoo Lee$^{2}$ \hspace{2mm}
Sohyeon Hong$^{1}$ \hspace{2mm}
Suhyun Jeon$^{1}$ \hspace{2mm}
Minkyu Kim$^{1}$ \\
\hspace*{1mm}\bfseries
Seungyeon Kim$^{1}$ \hspace{2mm}
Donghoon Han$^{3}$ \hspace{2mm}
Nojun Kwak$^{1}$ \\[0.5em]
$^{1}$Seoul National University \hspace{2mm}
$^{2}$Korea Advanced Institute of Science \& Technology \hspace{2mm}
$^{3}$Dnotitia
}

\newcommand{\algname}{CF-Prune}
\newcommand{\algfullname}{\textbf{C}omplementary \textbf{F}uture token \textbf{Prun}ing}

\iclrfinalcopy 
\begin{document}

\maketitle
\lhead{Preprint}

\begin{abstract}
World action models (WAMs) jointly predict future visual states and robot actions, but processing dense future-frame tokens incurs
substantial inference latency. We propose \textbf{\algname{}}, a training-free framework for pruning future-frame tokens in pretrained WAMs.
Our key insight is that future visual predictions support action
generation and therefore need not preserve every visual detail
required for high-fidelity video synthesis. \algname{} combines
instruction and action attention to identify task-relevant tokens
during the initial denoising forward pass, even when future-frame
and action representations remain highly noisy. An entropy-based
phase identification mechanism further retains dense computation
during precision phases to mitigate pruning-induced performance
degradation. Experiments on RoboLab, RoboCasa, and real-world
manipulation tasks demonstrate consistent inference speedups.
At a 35\% pruning ratio, \algname{} achieves average speedups of
$1.19$--$1.27\times$ while improving task success rates.
At a 50\% pruning ratio, it achieves $1.29$--$1.33\times$ average
speedups on both simulation benchmarks, with overall success
rates within 1.0 percentage point of the baseline.
\end{abstract}

\section{Introduction}
\label{sec:introduction}

\begin{wrapfigure}{r}{0.34\textwidth}
    \vspace{-4mm}
    \centering
    \includegraphics[width=\linewidth]{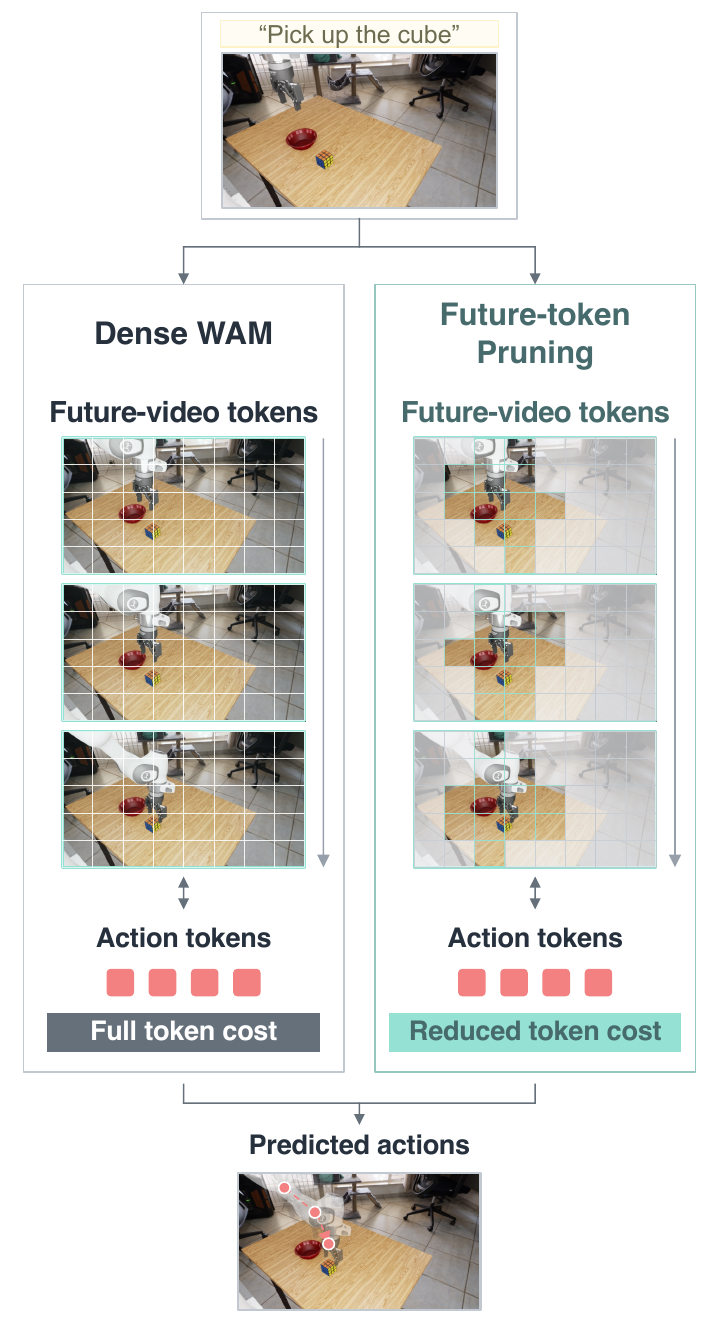}
    \caption{Rethinking future-video generation for action prediction.}
    \label{fig:conceptual}
    \vspace{-4mm}
\end{wrapfigure}

Extending foundation models to physical environments, particularly robotic manipulation, has recently attracted significant attention, due to their potential to enable robots to generalize across diverse tasks and environments. Vision-language-action (VLA) models represent a prominent approach \citep{brohan2023rt, kim2024openvla,black2024pi_0}, while world action models (WAMs) based on video diffusion models have emerged to better capture scene geometry, physical dynamics, and future state evolution \citep{kim2026cosmos,ye2026world}. However, deploying these models on real robots requires not only high task performance but also low-latency inference \citep{black2026real}. Since robotic manipulation requires repeated action generation throughout a task, inference latency accumulates across policy queries, delaying action execution and substantially increasing the overall task completion time. This problem is especially critical for WAMs due to the substantial computational overhead of video diffusion models, making efficient inference essential for their practical use in real-world manipulation. 

WAMs jointly generate robot actions and multiple future frames, introducing dense spatiotemporal token grids that account for a substantial portion of the computation in the video diffusion backbone. While VLAs address a similar bottleneck by pruning observation tokens \citep{yang2026efficientvla,li2025sp,wang2025specprune}, directly pruning future-frame tokens is difficult because they constitute the model outputs in conventional video diffusion models and support action generation through future-state prediction in WAMs. Video diffusion acceleration therefore modifies attention masks  \citep{xi2025sparse,xia2025training}, but the cost of determining effective masks can offset the gains in few-step WAMs. Existing WAM methods instead reduce denoising steps \citep{ye2026world} or redesign the model to bypass future generation \citep{yuan2026fast}; the former retains the per-step token burden, whereas the latter requires retraining. Consequently, existing WAM acceleration methods do not directly reduce future-frame tokens without additional training.

We propose \textbf{\algname{}} (\algfullname{}), a training-free method for pruning future-frame tokens in WAMs. Our key insight is that, while video diffusion models are designed to generate high-fidelity videos, WAMs use future-frame prediction primarily as an intermediate reference for action generation. Therefore, as
illustrated in Figure~\ref{fig:conceptual}, future representations must preserve task-relevant semantics and dynamics but need not retain every fine-grained visual detail. Allocating a dense grid of tokens to future frames is thus unnecessarily redundant for action prediction, motivating selective pruning that retains only the tokens needed for action generation without retraining. Realizing this potential requires identifying which future-frame tokens matter for action prediction early in denoising, when their representations remain highly noisy and their relevance is not yet explicit.

Specifically, \algname{} constructs an attention-based pruning criterion by analyzing the concentration of visual information with respect to action and instruction tokens, enabling effective token selection even when the action and future-frame latents remain highly noisy. Using only the initial forward pass, \algname{} identifies redundant tokens early in the denoising process, thereby maximizing pruning efficiency while demonstrating that the internal representations available at the beginning of generation are sufficient to select future-frame tokens needed for action generation. Finally, \algname{} introduces a phase identification mechanism that exploits entropy differences across camera views and temporal positions within the action sequence to identify precision phases and selectively relax pruning, mitigating performance degradation. 
Across RoboLab \citep{yang2026robolab}, RoboCasa \citep{nasiriany2024robocasa}, and real-world tasks, \algname{} achieves consistent efficiency gains at a 35\% pruning ratio: a $1.23\times$ speedup with a 1.3 percentage point success rate gain on RoboLab, a $1.19\times$ speedup with a 4.0 percentage point gain on RoboCasa, and a $1.27\times$ speedup with a 5.4 percentage point gain in real-world experiments. At a 50\% pruning ratio, \algname{} achieves at least a $1.29\times$ average inference speedup across both simulation benchmarks, while keeping success rates within 1.0 percentage point of the dense baseline.

Our main contributions are summarized as follows:
\begin{itemize}
    \item We reconsider the auxiliary role of future frames in WAMs and formulate token pruning from the perspective of action prediction, exposing redundancy that is not considered in conventional video generation.

    \item We propose \algname{}, a framework requiring no additional training that identifies redundant future frame tokens from noisy initial latents using their information concentration with respect to action and instruction tokens, while selectively relaxing pruning based on cross-view and temporal attention-entropy contrasts.
    
    \item We extensively evaluate \algname{} on the RoboLab and RoboCasa simulation benchmarks and real-world manipulation tasks, demonstrating consistent improvements in inference efficiency while preserving manipulation performance.
    
\end{itemize}
\section{Related Work}
\label{sec:relatedworks}
\paragraph{World Action Models}
World action models build upon video diffusion models to jointly predict robot actions and future visual states, transferring learned physical dynamics and predictive capabilities to robot control \citep{kim2026cosmos,ye2026world,agarwal2026cosmos}. Existing approaches either adapt pretrained video diffusion models through post-training on robot demonstrations \citep{kim2026cosmos,ye2026world} or jointly denoise video, action, and other multimodal tokens within a unified generator \citep{agarwal2026cosmos}. Other methods predict compact latent future representations or decouple video and action decoding for efficient action inference \citep{zheng2025flare,li2025unifiedvideoactionmodel}. More recent action-centered designs jointly train action prediction and future video generation \citep{ye2026gigaworld,yuan2026fast}. Across these approaches, future-state prediction serves as a key source of supervision or predictive context for action generation.

\paragraph{WAM Acceleration}
Although future prediction is a defining advantage of WAMs, it comes at substantial inference cost because generating numerous future-frame tokens significantly increases computation within the video diffusion backbone. Existing approaches reduce inference cost by using fewer denoising steps \citep{agarwal2026cosmos} or enabling single-step inference through decoupled video and action noise schedules during training \citep{ye2026world}. However, these methods accelerate inference by reducing denoising steps rather than making each step faster. Other approaches replace explicit future-frame prediction with compact future-state representations \citep{zheng2025flare} or avoid explicit future-video generation during inference \citep{ye2026gigaworld,li2025unifiedvideoactionmodel,yuan2026fast}. These methods require changes to the model architecture or training objective, making them difficult to apply directly to existing pretrained WAMs without additional training. In contrast, our approach directly reduces the future-frame token burden through token pruning, with no further training required.

\paragraph{VLA Token Pruning}
Vision-language-action (VLA) models typically combine a pretrained VLM backbone with an action prediction module to generate robot actions from visual observations and language instructions. Because processing dense visual tokens through large VLM backbones incurs substantial inference costs, recent studies have adapted visual token pruning techniques \citep{chen2024image,zhang2024sparsevlm} to VLA policies. Some methods select task-relevant and diverse visual tokens \citep{yang2026efficientvla} or jointly consider spatial and semantic importance \citep{li2025sp}, while others adjust pruning intensity across manipulation phases using recent robot motion \citep{pei2026action}. Further approaches combine semantic and action-decoding attention \citep{liu2025bridging}, incorporate historical context into current token selection \citep{wang2025specprune}, or forecast deep-layer token importance from attention patterns stored at previous control steps \citep{ma2026safe}. Overall, these methods extend VLM-based pruning to robotic policies, but are not directly applicable to noisy future-frame tokens, limiting their efficiency gains in WAMs where noisy future-frame tokens dominate computation.

\section{method}
\label{sec:method}
\begin{figure*}[t]
    \centering
    \includegraphics[width=\textwidth, trim={0 0 0 0}, clip]
    {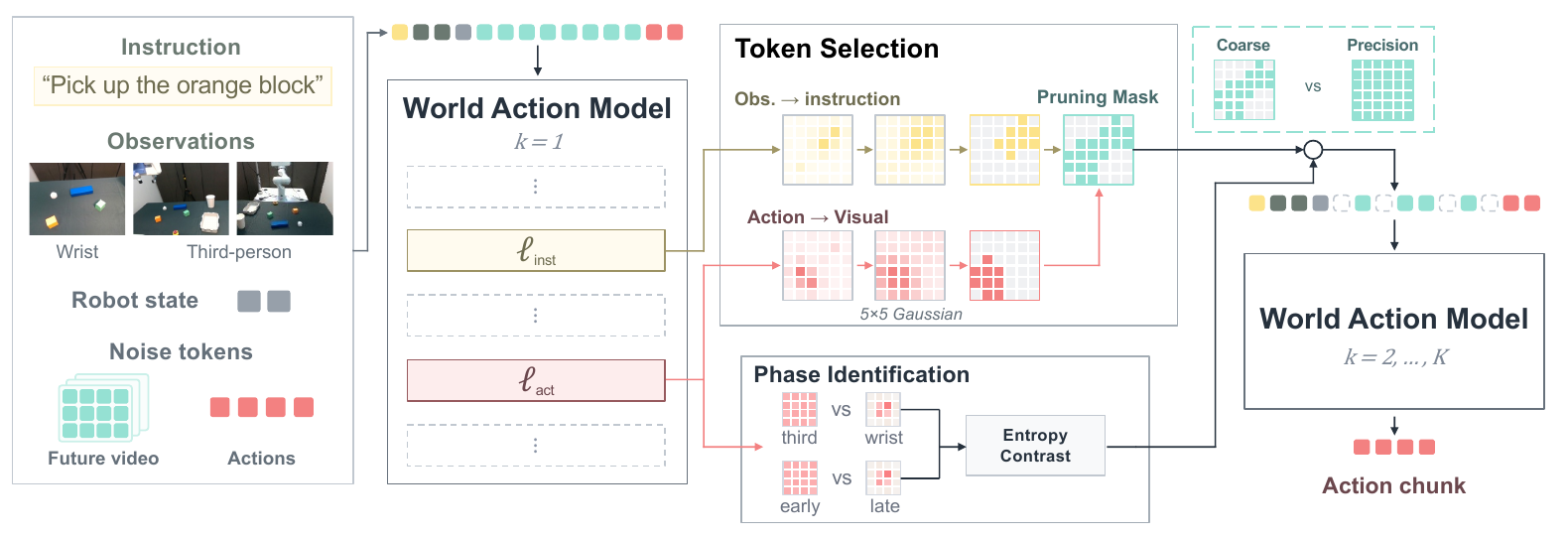}
    \caption{
    Overview of \algname{}.
    $k$ denotes the current denoising step, and $K$ is the total
    number of denoising steps.}
    \label{fig:architecture}
\end{figure*}
\subsection{Problem Formulation}
\label{sec:problem_formulation}
\paragraph{WAM Architecture.}
Video-generative world action models extend video diffusion models with an action modality to jointly predict future visual states and robot action trajectories. Given a language instruction, visual observations, and the current robot state, a WAM denoises future-video and action representations through either modality-specific generation experts or a unified diffusion transformer \citep{ye2026world,bi2026motus,agarwal2026cosmos}. The action prediction horizon $H_a$ denotes the number of future control steps generated by a single policy query. Since the complete action vector at each step is represented by one token, an action trajectory contains $H_a$ tokens regardless of the action dimensionality. In contrast, a future video is represented in the video VAE latent space as a spatiotemporal grid containing $T_v$ latent timesteps and $H_v\times W_v$ spatial positions per timestep. Flattening this grid produces $N_v=T_vH_vW_v$ future-video tokens.

\paragraph{Challenges.}
Even after spatiotemporal VAE compression, future-video tokens substantially outnumber action tokens, with $N_v \gg H_a$, and dominate computation at every denoising step. Directly pruning these tokens, however, is challenging because they constitute the model output rather than observed visual inputs and are gradually denoised from Gaussian noise. Estimating their importance only after sufficient visual information emerges reduces the acceleration benefit, whereas pruning them too early may discard information required for accurate action prediction. We therefore aim to identify redundant future-video tokens early in denoising without further training, reducing inference cost while preserving manipulation performance.

\subsection{Attention-Based Future-Frame Token Pruning}
\label{sec:attention_pruning}
We compute instruction and action relevance scores and complete
token selection within the initial denoising forward pass,
applying pruning throughout the remaining denoising process.
An overview of \algname{} is shown in Figure~\ref{fig:architecture}.
Observation-to-instruction attention provides a spatial prior
for instruction-relevant regions in the observed scene, while
action-to-visual attention, where visual tokens refer to both
observation and future-frame tokens, measures how strongly the
action tokens weight each visual region during generation.
We rank tokens using these two signals and combine the selected
token sets at a pruning ratio of $\alpha$.

\paragraph{Instruction-Based Spatial Prior.}
We use observation-to-instruction attention to identify spatial regions relevant to the instruction in the current observation. During the initial denoising step, action and future-frame tokens remain highly noisy, making their spatial information difficult to discern, whereas observation tokens represent the current scene without added diffusion noise. We therefore use observations as a spatial prior for instruction-relevant regions. To select a DiT layer for pruning, we normalize the instruction relevance scores across observation tokens at each layer and compute their entropy. This entropy measures the spatial concentration of instruction relevance scores, with lower values indicating that these scores are concentrated on fewer observation tokens.

Let $A_{q,k}^{(\ell,h)}$ denote the attention weight from query $q$ to key $k$ in head $h$ of layer $\ell$. Here, $x_j$ denotes the $j$-th instruction token among $N_{\mathrm{inst}}$ instruction tokens, and $o_i$ denotes the $i$-th observation token among the $N_o$ observation tokens across all camera views. The instruction relevance score of observation token $o_i$ at layer $\ell$ is defined as
\begin{equation}
s^{\mathrm{inst},\ell}(i)
=
\frac{1}{N_hN_{\mathrm{inst}}}
\sum_{h=1}^{N_h}
\sum_{j=1}^{N_{\mathrm{inst}}}
A_{o_i,x_j}^{(\ell,h)},
\label{eq:instruction_score}
\end{equation}
where $N_h$ is the number of attention heads.
We normalize the scores over the $N_o$ observation tokens and
compute the instruction attention entropy
$\mathcal{H}^{\mathrm{inst},\ell}$ at layer $\ell$ as follows:
\begin{equation}
\mathcal{H}^{\mathrm{inst},\ell}
=
-\frac{1}{\log N_o}
\sum_{i=1}^{N_o}
p^{\mathrm{inst},\ell}(i)
\log p^{\mathrm{inst},\ell}(i),
\qquad
p^{\mathrm{inst},\ell}(i)
=
\frac{s^{\mathrm{inst},\ell}(i)}
{\sum_{u=1}^{N_o}s^{\mathrm{inst},\ell}(u)}.
\label{eq:instruction_entropy}
\end{equation}
Here, we define $0\log 0=0$, yielding
$\mathcal{H}^{\mathrm{inst},\ell}\in[0,1]$. We measure this entropy during the initial denoising forward pass and select the layer with the lowest entropy.

\paragraph{Action-Based Visual Relevance.}
We complement the instruction-based information with the concentration of action-to-visual attention scores for pruning. Since the model ultimately aims to generate accurate actions, instruction relevance alone may not fully capture the visual information needed for action prediction. We therefore examine how strongly action tokens weight visual regions while accounting for differences in visual change across camera views. Observations and future frames comprise wrist and third-person views. The wrist camera moves with the robot, resulting in substantial visual changes across future frames. In contrast, fixed third-person views preserve the background, with changes primarily confined to robot and object motion. Accordingly, we use action-to-future attention scores for each future frame in the wrist view, while applying action-to-observation attention scores uniformly across future frames in the third-person views.

Here, $a_j$ denotes the $j$-th predicted action token, and $N_a$ is the number of action tokens. The index $i$ spans spatial positions across all camera views, and $f_{t,i}$ denotes the future-frame token at latent timestep $t$ corresponding to observation token $o_i$. $\mathcal{S}_{\mathrm{wrist}}$ and $\mathcal{S}_{\mathrm{third}}$ denote the sets of spatial indices belonging to wrist and third-person views, respectively.

For wrist views, we compute action relevance scores separately
for each future frame:
\begin{equation}
s_t^{\mathrm{act},\ell}(i)
=
\frac{1}{N_hN_a}
\sum_{h=1}^{N_h}
\sum_{j=1}^{N_a}
A_{a_j,f_{t,i}}^{(\ell,h)},
\qquad i\in\mathcal{S}_{\mathrm{wrist}}.
\label{eq:action_future_score}
\end{equation}
For third-person views, we use action relevance scores shared
across all future frames:
\begin{equation}
s_t^{\mathrm{act},\ell}(i)
=
\frac{1}{N_hN_a}
\sum_{h=1}^{N_h}
\sum_{j=1}^{N_a}
A_{a_j,o_i}^{(\ell,h)},
\qquad i\in\mathcal{S}_{\mathrm{third}},\quad \forall t.
\label{eq:action_observation_score}
\end{equation}

For layer selection, we compute the action attention entropy
$\mathcal{H}^{\mathrm{act},\ell}$ analogously to
Eq.~(\ref{eq:instruction_entropy}). Specifically, we average
action-to-future attention over heads and action tokens, normalize
the resulting scores over all future-frame tokens across camera
views and time steps, and compute their normalized entropy. The full formulation is provided in
Appendix~\ref{app:action_entropy}.

\paragraph{Layer Selection.}
We use the entropy criteria defined above to select layers for instruction and action attention scoring based on the initial denoising step. Applying these criteria independently at every policy query would require evaluating multiple layers during the initial denoising forward pass. However, layer-wise entropy profiles remain similar across tasks and rollout stages, with low entropy consistently occurring in similar layers. We therefore select the scoring layers, $\ell_{\mathrm{inst}}$ and $\ell_{\mathrm{act}}$, at the first policy query in each environment and keep them fixed thereafter to avoid repeated layer selection during inference. The supporting analysis and specific layer choices are presented in Section~\ref{sec:analysis}. The resulting relevance scores $s^{\mathrm{inst},\ell_{\mathrm{inst}}}$ and $s^{\mathrm{act},\ell_{\mathrm{act}}}$ are used for token selection.

\paragraph{Spatial Smoothing.}
Attention scores often concentrate on a small subset of object patches, causing attention-based token selection to discard other parts of the same object. Inspired by the spatial smoothing in EADP \citep{wang2026combating}, we reshape the aggregated attention scores into 2D spatial maps and apply a $5\times5$ Gaussian blur before token selection. This smoothing spreads relevance to neighboring patches to encourage spatially contiguous selections. A qualitative comparison before and after smoothing is provided in Section~\ref{sec:analysis}.

\paragraph{Action--Instruction Complementary Selection.}
Using the relevance scores extracted from the selected layers, $s^{\mathrm{inst},\ell_{inst}}$ and $s^{\mathrm{act},\ell_{act}}$, we perform complementary token selection based on action and instruction relevance. Given a pruning ratio $\alpha$, we first retain the top $(1-\alpha)/2$ fraction of tokens according to $s^{\mathrm{act},\ell_{act}}$ to preserve the visual regions most relevant to action generation. We then exclude the already selected tokens and retain an additional $(1-\alpha)/2$ fraction according to $s^{\mathrm{inst},\ell_{inst}}$. This procedure preserves a total fraction of $1-\alpha$ tokens while incorporating complementary information from both relevance signals. For the wrist view, token selection is performed independently for each future frame. For the third-person view, both relevance scores are derived from the observation tokens; therefore, a shared token selection is applied across all future frames.

\subsection{Entropy-Based Phase Identification}
\label{sec:precision_phase}

We identify manipulation phases requiring precise object interaction as precision phases and detect them using the attention entropy of current action tokens during the initial denoising forward pass. Although previously generated action chunks provide motion cues for phase estimation, they may respond slowly to phase transitions because WAMs use long action horizons. We define the entropy contrast score as $C=C_{\mathrm{view}}+C_{\mathrm{temp}}$, where $C_{\mathrm{view}}=\mathcal{H}_{\mathrm{third}}-\mathcal{H}_{\mathrm{wrist}}$ measures the relative concentration of action-to-observation attention near the gripper, and $C_{\mathrm{temp}}=\mathcal{H}_{\mathrm{early}}-\mathcal{H}_{\mathrm{late}}$ captures increasing localization of action-to-future attention toward later actions. Together, these two terms indicate transitions toward localized visual dependencies associated with precise object interaction. We classify an inference call as a precision phase when $C>\tau$ and as a coarse phase otherwise. Pruning is disabled during precision phases and applied only during coarse phases. Details of the phase entropy computation are provided in
Appendix~\ref{app:phase_id}.
\section{Experiments}

\subsection{Experimental Settings}
\label{sec:experimental_setup}
\begin{figure*}[t]
    \centering
    \captionsetup[subfigure]{
        justification=centering,
        singlelinecheck=false
    }

    \begin{subfigure}[t]{0.24\textwidth}
        \centering
        \includegraphics[width=\linewidth]{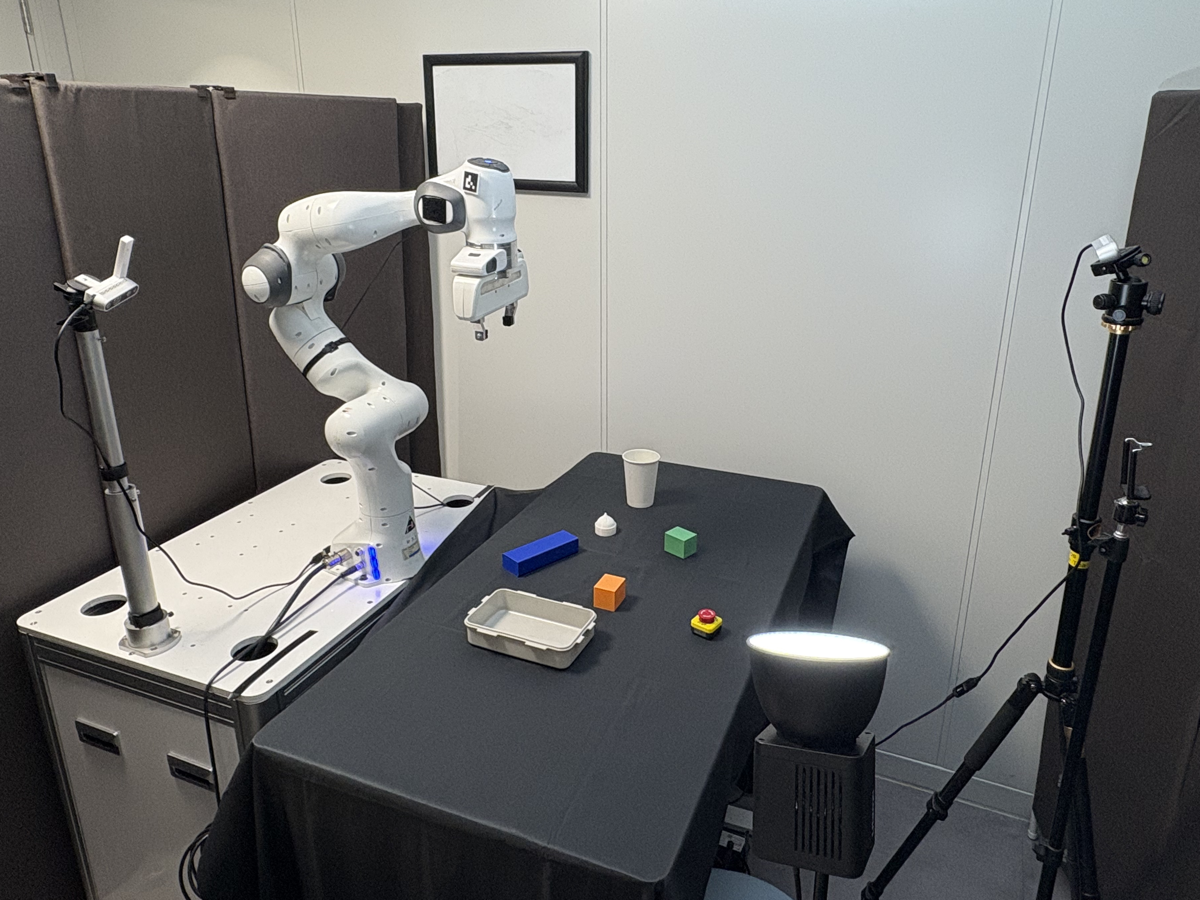}
        \caption{Experimental setup}
        \label{fig:real_setup}
    \end{subfigure}\hfill
    \begin{subfigure}[t]{0.24\textwidth}
        \centering
        \includegraphics[width=\linewidth]{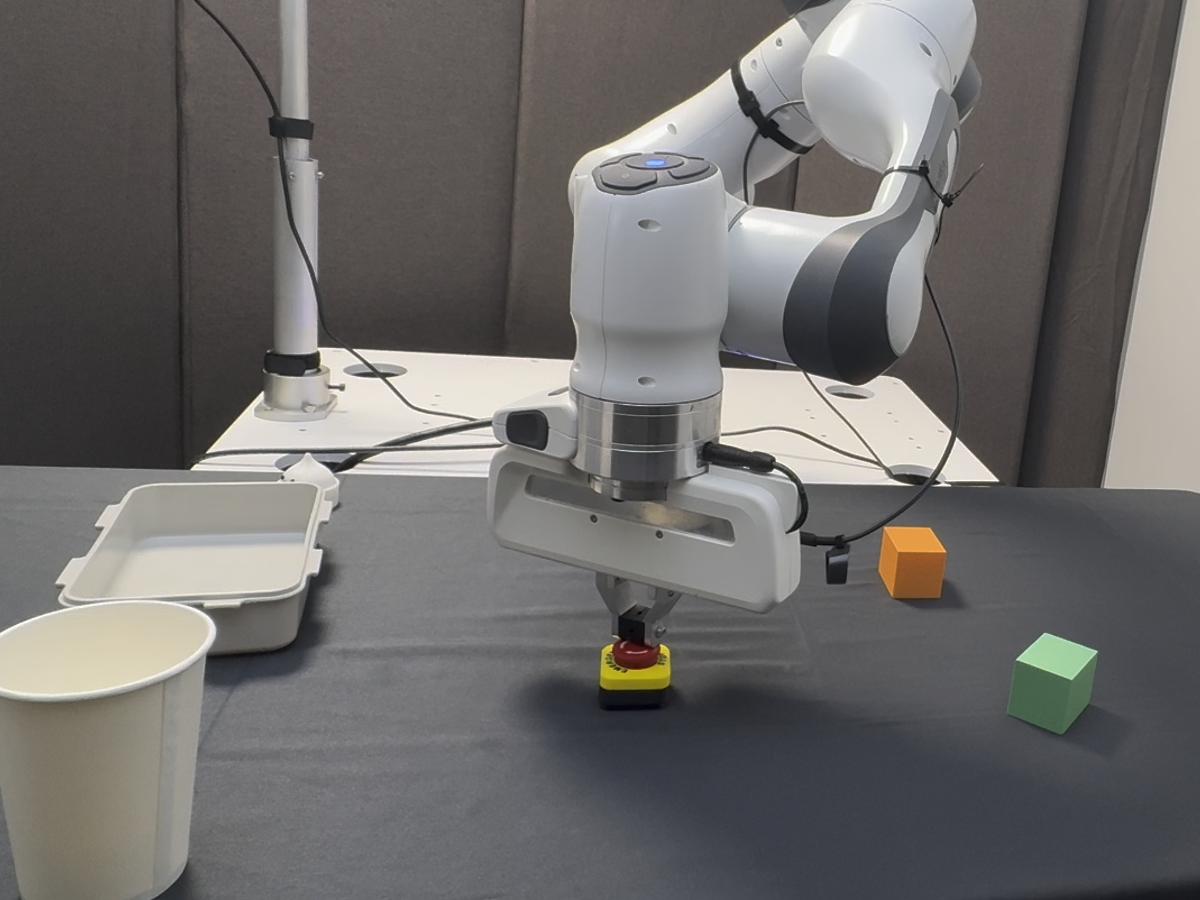}
        \caption{E-button pressing}
        \label{fig:real_button}
    \end{subfigure}\hfill
    \begin{subfigure}[t]{0.24\textwidth}
        \centering
        \includegraphics[width=\linewidth]{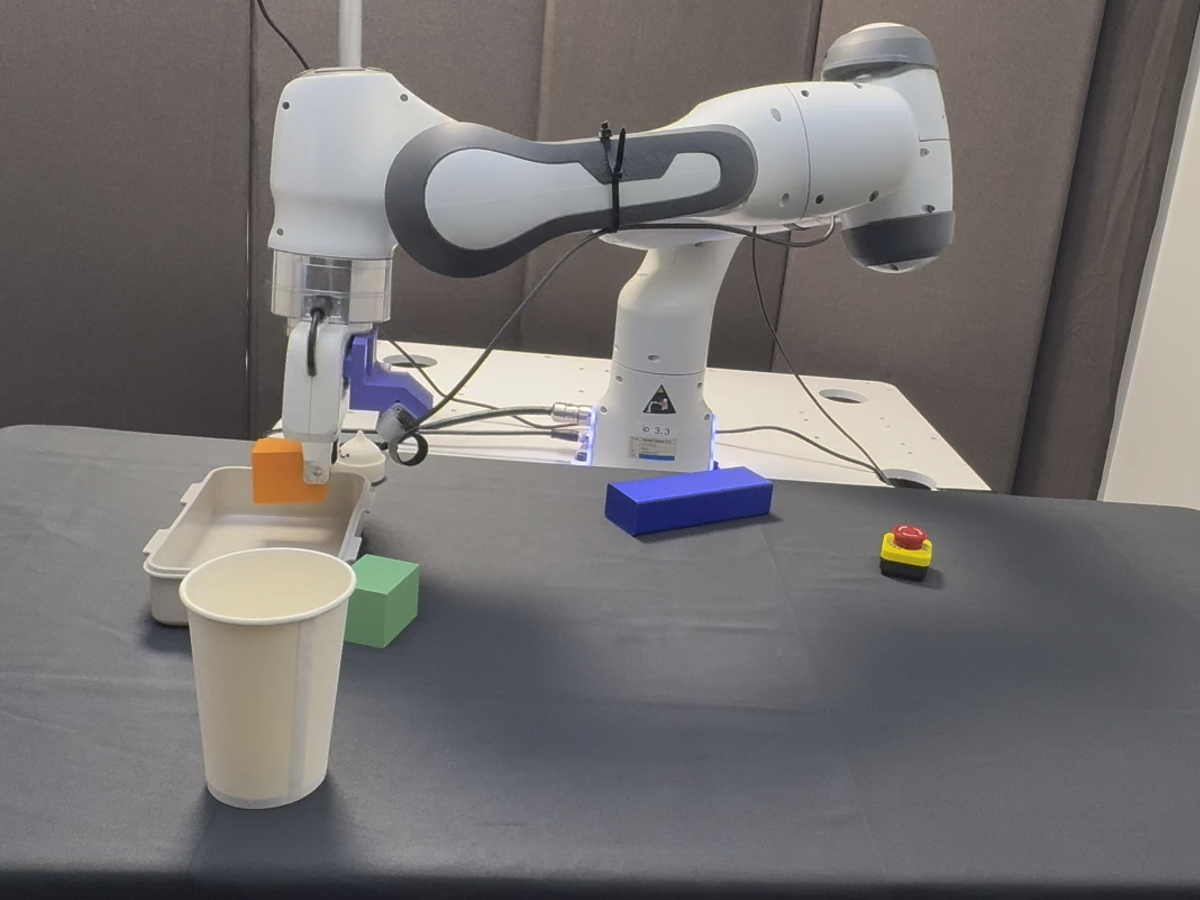}
        \caption{Block pick-and-place}
        \label{fig:real_pick}
    \end{subfigure}\hfill
    \begin{subfigure}[t]{0.24\textwidth}
        \centering
        \includegraphics[width=\linewidth]{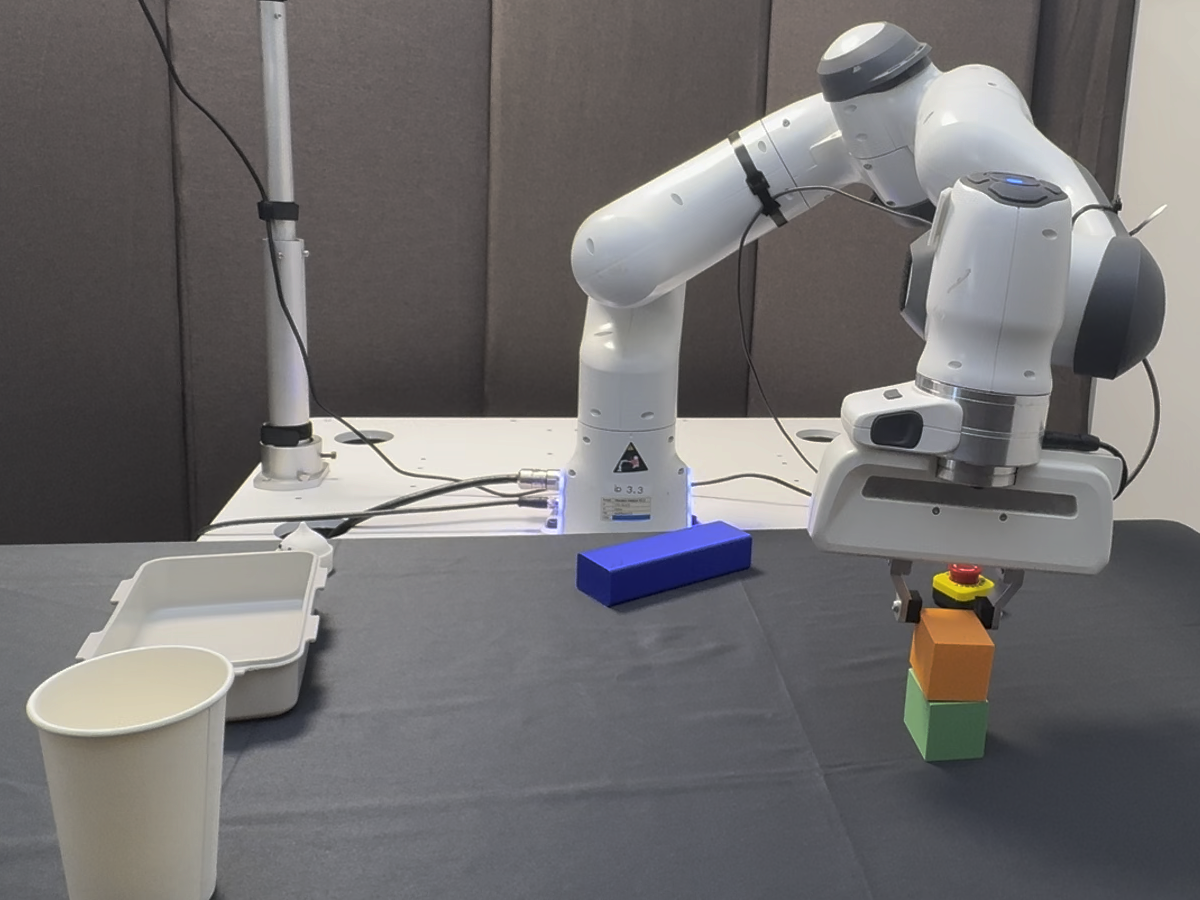}
        \caption{Block stacking}
        \label{fig:real_stack}
    \end{subfigure}

    \caption{
        Real-world experimental setup and the three manipulation tasks
        used for evaluation.
    }
    \label{fig:realworld_setup}
\end{figure*}
\paragraph{Benchmarks.}
We evaluate our method on RoboLab \citep{yang2026robolab} and RoboCasa \citep{nasiriany2024robocasa}. RoboLab comprises 64 Simple, 39 Moderate, and 17 Complex manipulation tasks. We perform 10 rollouts per task to measure success rates. Each task provides three instruction variants with different levels of specificity: Vague, Default, and Specific. We use the same Default instructions for all compared methods. On RoboCasa, we select one representative task from each category and evaluate each task over 100 rollouts.

\paragraph{Real-World Setup.}
We further evaluate \algname{} on a Franka Emika Panda platform
using three tabletop manipulation tasks: pressing an emergency
button, picking and placing a block, and stacking one block on
top of another. Objects are randomly positioned within a
$50 \times 50\,\mathrm{cm}$ region on the tabletop.
The policy receives observations from three cameras: a
wrist-mounted camera and two fixed cameras providing right
and front views of the workspace. The two fixed-camera views
serve as third-person observations.
Figure~\ref{fig:realworld_setup} shows the experimental setup
and the three manipulation tasks. Representative images from each camera view are shown in
Appendix~\ref{app:camera_setup}. We use the same camera and action configurations during
training and evaluation. Each method is evaluated over
25 rollouts per task using the same fine-tuned checkpoint.

\paragraph{Model and Fine-Tuning.}
We use Cosmos3 as the base world action model and adopt identical checkpoints and input configurations for all compared methods within each benchmark. For RoboLab, we use the pretrained Cosmos3-Nano-Policy-DROID model, trained on DROID \citep{khazatsky2024droid}, without additional training. A full-size wrist view is placed above two half-size third-person views arranged side by side, forming a single observation image. For RoboCasa, we fine-tune Cosmos3-Nano for 16,000 steps with a batch size of 1,024. One wrist view and one third-person view are concatenated side by side as the visual observation. For real-world experiments, we collect 60 successful demonstrations per task and fine-tune Cosmos3-Nano-Policy-DROID using LoRA \citep{hu2021lora}. The three camera views are combined into a single observation image using the same layout as in RoboLab. Token pruning is applied only during inference and requires no additional model training. Additional training and inference settings are provided in
Appendix~\ref{app:implementation_details}.

\begin{table*}[t]
\centering
\footnotesize
\setlength{\tabcolsep}{3.5pt}
\renewcommand{\arraystretch}{1.1}
\caption{
Results on RoboLab. Success rates are averaged over tasks.
Coarse phase latency excludes precision phases, while average
latency and FLOPs are averaged over all inference calls.
}
\label{tab:robolab_results}

\begin{tabular}{ccccccccc}
\toprule
\multirow{2}{*}{Method}
& \multirow{2}{*}{Variant}
& \multicolumn{4}{c}{Success Rate (\%)}
& \multicolumn{2}{c}{Speedup}
& \multirow{2}{*}{FLOPs} \\
\cmidrule(lr){3-6}
\cmidrule(lr){7-8}
& & Simple & Moderate & Complex & Overall
& Coarse & Average & \\
\midrule
Dense
& --
& 40.3 & 34.1 & 27.6 & 36.5
& -- & $1.00\times$ & 100\% \\
\midrule
\multirow{2}{*}{SpecPrune-VLA}
& Obs-only (70\%)
& 33.4 & 33.8 & 24.7 & 32.3
& $1.07\times$ & $1.06\times$ & 93.1\% \\
& Obs + future (25\%)
& 36.9 & 28.2 & 23.5 & 32.2
& $1.26\times$ & $1.23\times$ & 77.5\% \\
\midrule
\multirow{2}{*}{SP-VLA}
& Obs-only (70\%)
& 34.4 & 24.1 & 21.2 & 29.2
& $1.05\times$ & $1.05\times$ & 93.1\% \\
& Obs + future (25\%)
& 37.5 & 32.3 & 23.5 & 33.8
& $1.24\times$ & $1.24\times$ & 77.5\% \\
\midrule
Sparse VideoGen
& --
& 21.9 & 16.9 & 7.1 & 18.2
& -- & $0.97\times$ & 104.7\% \\
\midrule
\rowcolor{oursbg}
& $\alpha=35\%$
& 43.0 & 33.6 & 28.2 & 37.8
& $1.37\times$ & $1.23\times$ & 73.2\% \\
\rowcolor{oursbg}
\multirow{-2}{*}{\algname{} (Ours)}
& $\alpha=50\%$
& 40.3 & 32.8 & 24.7 & 35.7
& $1.57\times$ & $1.33\times$ & 61.7\% \\
\bottomrule
\end{tabular}
\end{table*}
\begin{table*}[t]
\centering
\footnotesize
\setlength{\tabcolsep}{3pt}
\renewcommand{\arraystretch}{1.1}
\caption{
Results on four representative RoboCasa tasks, with 100 rollouts per task.
The columns represent articulation (OpenStandMixerHead),
object transfer (CoffeeSetupMug), device control (TurnOnElectricKettle),
and navigation (NavigateKitchen).
}
\label{tab:robocasa_results}

\begin{tabular}{cccccccccc}
\toprule
\multirow{2}{*}{Method}
& \multirow{2}{*}{Variant}
& \multicolumn{5}{c}{Success Rate (\%)}
& \multicolumn{2}{c}{Speedup}
& \multirow{2}{*}{FLOPs} \\
\cmidrule(lr){3-7}
\cmidrule(lr){8-9}
& & Artic. & Transfer & Control & Nav. & Overall
& Coarse & Average & \\
\midrule
Dense
& --
& 95.0 & 55.0 & 72.0 & 38.0 & 65.0
& -- & $1.00\times$ & 100\% \\
\midrule
\multirow{2}{*}{SpecPrune-VLA}
& Obs-only (70\%)
& 65.0 & 17.0 & 72.0 & 44.0 & 49.5
& $1.06\times$ & $1.05\times$ & 95.4\% \\
& Obs + future (25\%)
& 92.0 & 51.0 & 78.0 & 36.0 & 64.3
& $1.21\times$ & $1.17\times$ & 82.8\% \\
\midrule
\multirow{2}{*}{SP-VLA}
& Obs-only (70\%)
& 82.0 & 27.0 & 75.0 & 42.0 & 56.5
& $1.07\times$ & $1.07\times$ & 93.7\% \\
& Obs + future (25\%)
& 94.0 & 38.0 & 90.0 & 37.0 & 64.8
& $1.18\times$ & $1.17\times$ & 82.8\% \\
\midrule
Sparse VideoGen
& --
& 92.0 & 55.0 & 77.0 & 37.0 & 65.3
& -- & $1.01\times$ & 99.3\% \\
\midrule
\rowcolor{oursbg}
& $\alpha=35\%$
& 97.0 & 68.0 & 76.0 & 35.0 & 69.0
& $1.29\times$ & $1.19\times$ & 84.4\% \\
\rowcolor{oursbg}
\multirow{-2}{*}{\algname{} (Ours)}
& $\alpha=50\%$
& 94.0 & 71.0 & 73.0 & 18.0 & 64.0
& $1.47\times$ & $1.29\times$ & 77.7\% \\
\bottomrule
\end{tabular}
\end{table*}
\begin{table*}[t]
\centering
\footnotesize
\setlength{\tabcolsep}{6pt}
\renewcommand{\arraystretch}{1.1}

\caption{
Real-world evaluation results.
Success rates are measured over 25 rollouts per task, and Overall denotes
the average success rate across the three tasks.
}
\label{tab:realworld_results}

\begin{tabular}{ccccccc}
\toprule
\multirow{2}{*}{Method}
& \multicolumn{4}{c}{Success Rate (\%)}
& \multicolumn{2}{c}{Speedup} \\
\cmidrule(lr){2-5}
\cmidrule(lr){6-7}
& \shortstack{Emergency\\Button}
& \shortstack{Block\\Pick-and-Place}
& \shortstack{Block\\Stacking}
& Overall & Coarse & Average \\
\midrule
Dense
& 84.0 & 92.0 & 80.0 & 85.3
& -- & $1.00\times$ \\
\midrule
\rowcolor{oursbg}
\algname{} ($\alpha=35\%$)
& 88.0 & 96.0 & 88.0 & 90.7
& $1.37\times$ & $1.27\times$ \\
\bottomrule
\end{tabular}
\end{table*}
\paragraph{Compared Methods.}
We compare five inference methods: (1) a baseline without token pruning, (2) \algname{}, (3) SpecPrune-VLA \citep{wang2025specprune} adapted to WAM, (4) SP-VLA \citep{li2025sp} adapted to WAM, and (5) Sparse VideoGen \citep{xi2025sparse}, which sparsifies attention connections. We evaluate \algname{} at pruning ratios of 35\% and 50\%. For both SpecPrune-VLA and SP-VLA, we evaluate two variants: \emph{Obs-only}, which prunes 70\% of observation tokens, and \emph{Obs + future}, which selects a 25\% pruning mask from observation tokens and applies it to the corresponding spatial positions in both the observations and all future latent frames. The 25\% pruning ratio is chosen to yield average inference speedups comparable to those of our method at 35\% pruning, allowing manipulation performance to be compared at similar inference costs. Unlike token pruning, Sparse VideoGen retains all tokens but restricts the connections used in attention computation.

\paragraph{Evaluation Metrics.}
We report per-task success rates and average success rates for each difficulty level, together with inference latency and FLOPs reductions relative to the baseline. All latency measurements are performed on an NVIDIA RTX A6000 GPU under identical inference settings. We distinguish between inference speedup during coarse phases, where pruning is applied, and overall speedup accounting for the proportions of inference calls in precision and coarse phases observed during actual rollouts.

\subsection{Main Results}
\paragraph{RoboLab.}
As shown in Table~\ref{tab:robolab_results}, \algname{} improves inference efficiency while preserving manipulation performance. At $\alpha=35\%$, it achieves a $1.23\times$ average speedup and a 26.8\% FLOPs reduction, while increasing the overall success rate from 36.5\% to 37.8\%. This improvement suggests that selective pruning may regularize the policy by suppressing task-irrelevant visual information. At $\alpha=50\%$, \algname{} achieves a $1.33\times$ speedup and a 38.3\% FLOPs reduction with only a 0.8 percentage point decrease in success rate. Observation-only VLA pruning provides limited acceleration, whereas extending its masks to future frames causes larger performance degradation at comparable speedups. Sparse VideoGen is slower than the dense baseline because its profiling overhead outweighs its computational savings. Overall, \algname{} provides a favorable balance between manipulation performance and inference efficiency.

\paragraph{RoboCasa.}
As shown in Table~\ref{tab:robocasa_results}, at $\alpha=35\%$,
\algname{} achieves a $1.19\times$ average speedup and a 15.6\%
reduction in FLOPs, while improving the overall success rate
from 65.0\% to 69.0\%, consistent with the improvement observed
on RoboLab.
Observation-only VLA pruning provides limited acceleration
and substantially reduces overall success rates.
Extending pruning to future frames achieves comparable speedups
to our 35\% variant but lower overall success rates,
while Sparse VideoGen provides negligible acceleration.
Increasing our pruning ratio to $\alpha=50\%$ yields a
$1.29\times$ average speedup with an overall success rate of 64.0\%,
1.0 percentage point below the dense baseline.
However, the effect varies across tasks: articulation, transfer,
and control remain comparable to or better than the dense baseline,
whereas navigation success drops from 38.0\% to 18.0\%,
indicating greater sensitivity to aggressive pruning.

\paragraph{Real-World Evaluation.}
As shown in Table~\ref{tab:realworld_results}, \algname{} achieves
slightly higher success rates than the dense baseline across all
three real-world tasks. With a pruning ratio of $\alpha=35\%$,
\algname{} achieves success rates of 88.0\%, 96.0\%, and 88.0\%
on emergency-button pressing, block pick-and-place, and block
stacking, respectively, resulting in an overall success rate of
90.7\% compared with 85.3\% for the dense baseline.
It also provides a $1.37\times$ speedup during coarse phases
and an average inference speedup of $1.27\times$. These results demonstrate that future-frame token pruning
transfers effectively to real-world manipulation, reducing
inference latency without compromising task performance.
The success rate improvements at $\alpha=35\%$ across both
simulation benchmarks and real-world tasks further suggest
that suppressing task-irrelevant visual information may
benefit manipulation performance, as discussed in the
RoboLab evaluation.

\begin{wraptable}{r}{0.43\textwidth}
    \vspace{-13pt}
    \centering
    \footnotesize
    \setlength{\tabcolsep}{6pt}
    \renewcommand{\arraystretch}{1.15}
    \caption{
        Token selection strategy ablation on RoboLab
        at 35\% pruning.
    }
    \label{tab:ablation}
    \begin{tabular}{lc}
        \toprule
        Selection Strategy & Success Rate (\%) \\
        \midrule
        Instruction-only & 34.0 \\
        Action-only & 30.8 \\
        Instruction-first & 34.3 \\
        \midrule
        \rowcolor{oursbg}
        Action-first (Ours) & \textbf{35.3} \\
        \bottomrule
    \end{tabular}
\end{wraptable}

\subsection{Ablation}
\paragraph{Action--Instruction Complementary Selection.}
We determine which future-frame tokens to prune using action--instruction complementary selection. To evaluate its effectiveness, we compare our selection criterion with variants based solely on instruction or action relevance scores. We also examine the selection order by reversing our action-first strategy: selecting instruction-relevant tokens first, followed by action-relevant tokens from the remaining candidates. To isolate the contribution of the selection criterion, we fix the pruning ratio at 35\%, disable phase identification, and keep all other components unchanged across variants. All variants are evaluated across all 120 RoboLab tasks. As shown in Table~\ref{tab:ablation}, the instruction-only and action-only variants achieve overall success rates of 34.0\% and 30.8\%, respectively, whereas our complementary selection achieves 35.3\%. The instruction-first variant achieves 34.3\%, compared with 35.3\% for our action-first strategy. Combining action and instruction relevance yields higher success rates than either signal alone, with action-first selection outperforming instruction-first selection. A qualitative comparison of the selected regions is provided
in Appendix~\ref{app:selection_comparison}.

\begin{figure*}[t]
    \centering

    \begin{subfigure}[t]{0.49\textwidth}
        \centering
        \includegraphics[width=\linewidth]
        {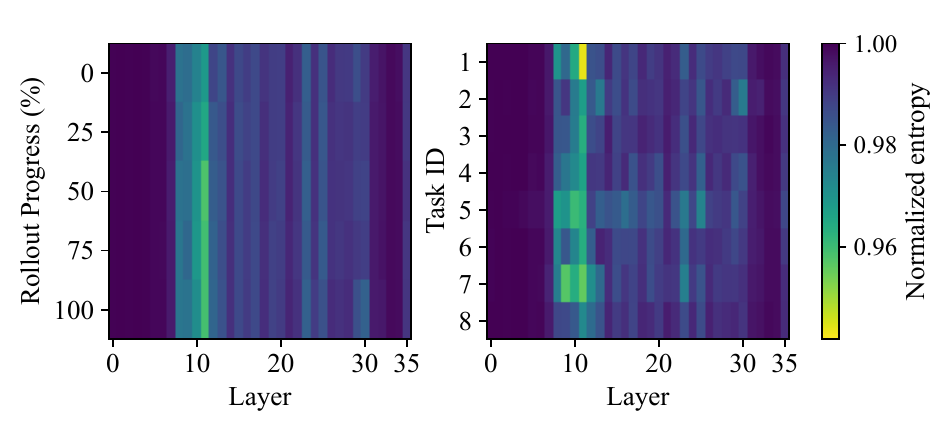}
        \caption{Instruction-based attention entropy}
        \label{fig:entropy_instruction}
    \end{subfigure}
    \hfill
    \begin{subfigure}[t]{0.49\textwidth}
        \centering
        \includegraphics[width=\linewidth]
        {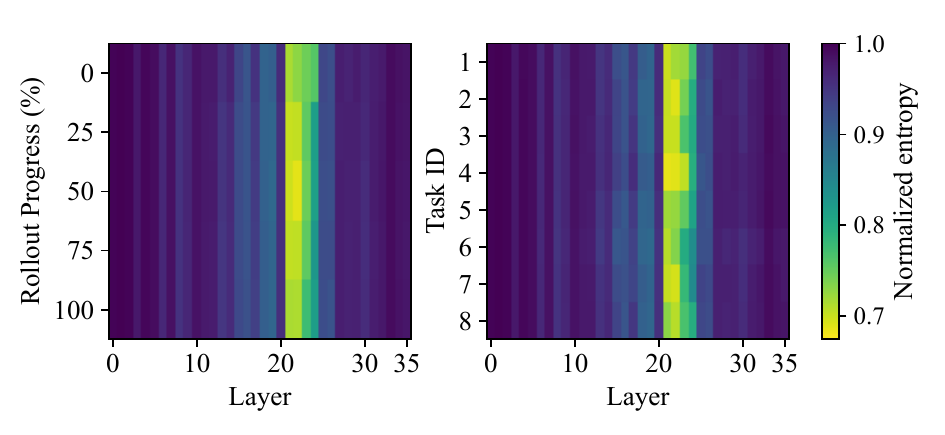}
        \caption{Action-based attention entropy}
        \label{fig:entropy_action}
    \end{subfigure}

    \caption{
Layer-wise normalized attention entropy at the first denoising
step across sampled RoboLab tasks.
In each subfigure, the left heatmap shows entropy over rollout
progress averaged across tasks, while the right shows
task-wise entropy averaged over sampled rollout stages.
}
    \label{fig:entropy}
\end{figure*}
\paragraph{Ablation on Entropy-Based Phase Identification.}
We evaluate the contribution of entropy-based phase identification by comparing fixed pruning against phase-adaptive pruning over all 120 RoboLab tasks. Both variants use the same 35\% token-pruning policy during coarse phases. The fixed variant applies pruning to every inference call, whereas the phase-adaptive variant retains dense computation during detected precision phases. Incorporating phase identification improves the overall success rate from 35.3\% to 37.8\%, while classifying only 31.4\% of inference calls as precision phases. Consequently, the method preserves an average speedup of 1.23$\times$. These results show that selectively restoring dense computation
during detected precision phases provides a favorable balance
between task performance and inference efficiency.

\subsection{Analysis}
\label{sec:analysis}
\paragraph{Layer-Wise Attention Entropy Consistency.}
Figure~\ref{fig:entropy} visualizes the normalized attention entropy across DiT layers during the initial denoising step over multiple tasks and rollout stages. Our entropy-based selection at the first policy query in each environment yields $\ell_{\mathrm{inst}}=11$ and $\ell_{\mathrm{act}}=23$, which are kept fixed thereafter. Instruction relevance is most concentrated around layer 11, whereas action relevance exhibits a consistently low entropy region across layers 21--23. These low-entropy regions remain stable across task instances
and rollout stages despite changes in scene configuration and manipulation phase, supporting the use of the initially selected layers without reselection.

\begin{figure*}[t]
    \centering
    \begin{subfigure}[t]{0.24\textwidth}
        \includegraphics[width=\linewidth]{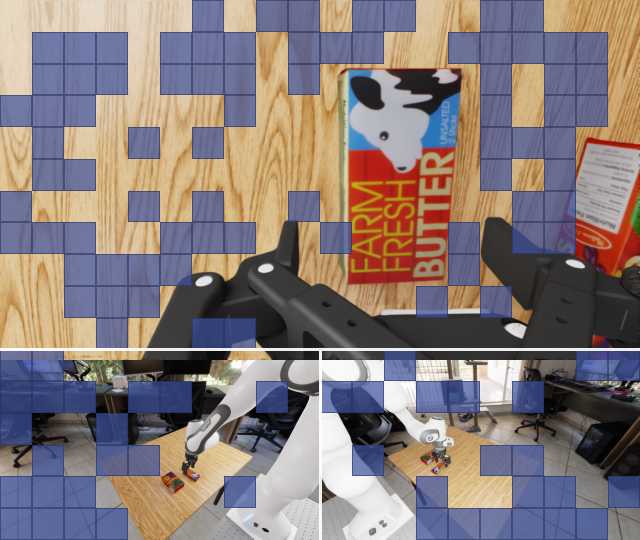}
        \caption{Raw, 35\%}
    \end{subfigure}
    \hfill
    \begin{subfigure}[t]{0.24\textwidth}
        \includegraphics[width=\linewidth]{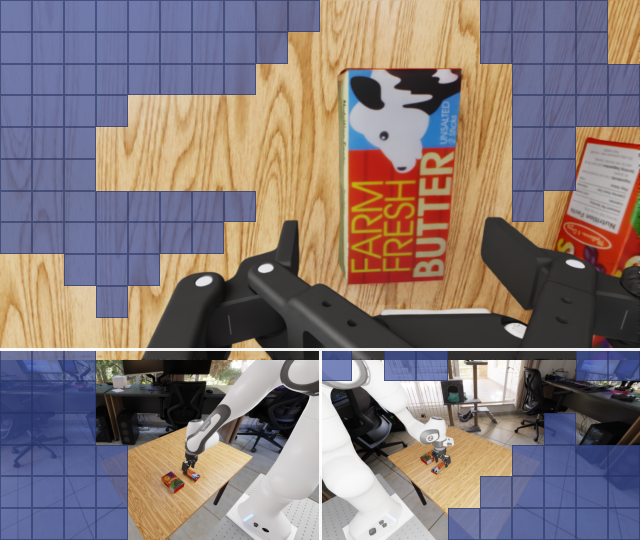}
        \caption{Smoothing, 35\%}
    \end{subfigure}
    \hfill
    \begin{subfigure}[t]{0.24\textwidth}
        \includegraphics[width=\linewidth]{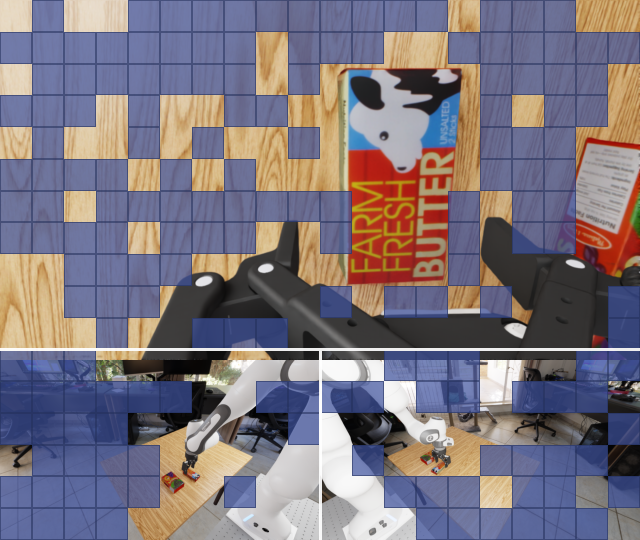}
        \caption{Raw, 50\%}
    \end{subfigure}
    \hfill
    \begin{subfigure}[t]{0.24\textwidth}
        \includegraphics[width=\linewidth]{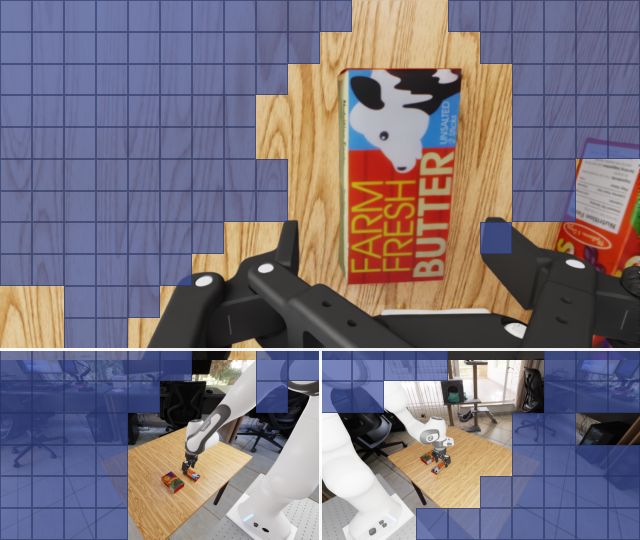}
        \caption{Smoothing, 50\%}
    \end{subfigure}
    \caption{
    Effect of Gaussian smoothing on token pruning.
    Navy cells mark pruned positions in the first future latent
    frame, overlaid on the input observation.
}
    \label{fig:blur}
\end{figure*}
 \paragraph{Effect of Spatial Smoothing.}
Raw relevance scores often concentrate on a few patches, causing token selection to discard other parts of relevant objects and their surrounding context. As shown in Figure~\ref{fig:blur}, Gaussian smoothing propagates relevance to neighboring patches, retaining a broader and more spatially coherent region around relevant objects. We evaluate smoothing on all 120 RoboLab tasks at 35\% pruning, keeping all other components of \algname{} unchanged. Gaussian smoothing improves the overall success rate from 36.0\% to 37.8\% with negligible computational overhead, demonstrating that preserving spatial context around relevant objects improves pruning robustness.
\section{Conclusion}
\label{sec:conclusion}
We presented \algname{}, a training-free framework for accelerating
pretrained WAMs through future-frame token pruning. By reconsidering
future visual prediction as an intermediate representation for action
generation, \algname{} combines instruction and action attention to
identify redundant tokens during the initial denoising forward pass.
Entropy-based phase identification further preserves dense computation
during precision phases. Across RoboLab, RoboCasa, and real-world
tasks, 35\% pruning achieves $1.19$--$1.27\times$ average inference
speedups while improving task success rates. Increasing the pruning
ratio to 50\% yields $1.29$--$1.33\times$ average speedups across
both simulation benchmarks, with success rates remaining within
1.0 percentage point of the dense baseline. These results demonstrate
that selective future-frame token pruning can reduce WAM inference
latency while preserving manipulation performance, without additional
training.

\subsection*{AI use statement}

We used generative AI tools to refine the academic phrasing and improve the flow of the manuscript.
We also used these tools to assist with broad searches for related work alongside our independent
literature search and to assist with writing code. The authors reviewed the AI-assisted writing, code,
and references and take responsibility for the final content of the paper.

\bibliography{iclr2027_conference}
\bibliographystyle{iclr2027_conference}

\clearpage
\appendix
\section{Appendix}

\subsection{Implementation Details}
\label{app:implementation_details}

For RoboLab, we use the pretrained Cosmos3-Nano-Policy-DROID
checkpoint without additional task-specific fine-tuning.
For RoboCasa, we fine-tune Cosmos3-Nano for 16,000 updates
with an effective batch size of 1,024 and a base learning rate
of $3.5\times10^{-5}$, using an action prediction horizon of 32.
For real-world experiments, we adapt Cosmos3-Nano-Policy-DROID
using LoRA with rank 32 and scaling parameter 32, applied to
the query, key, value, and output projections of the generation
expert.
We fine-tune each real-world policy for five epochs on
task-specific demonstrations using AdamW with a learning rate
of $5\times10^{-5}$, zero weight decay, and BF16 precision.

We follow the default inference settings of the respective
models, using four denoising steps for RoboLab and real-world
experiments and ten for RoboCasa.
These settings are kept unchanged across the dense baseline
and all compared pruning methods. The precision-phase thresholds are 0.05,
-0.15, and 0.15 for RoboLab, RoboCasa, and real-world
experiments, respectively.

\subsection{Real-world Camera Setup}
\label{app:camera_setup}
\begin{figure*}[htbp]
    \centering
    \begin{subfigure}[t]{0.32\textwidth}
    \centering
    \includegraphics[
        width=\linewidth,
        angle=180,
        origin=c
    ]{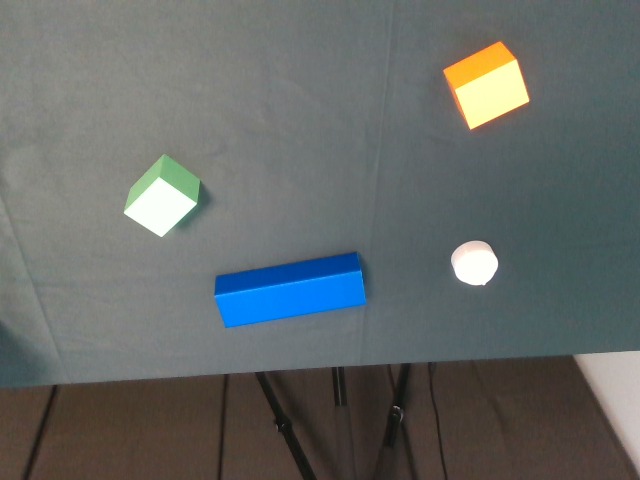}
    \caption{Wrist View}
    \label{fig:realworld_wrist}
\end{subfigure}
    \hfill
    \begin{subfigure}[t]{0.32\textwidth}
        \centering
        \includegraphics[width=\linewidth]{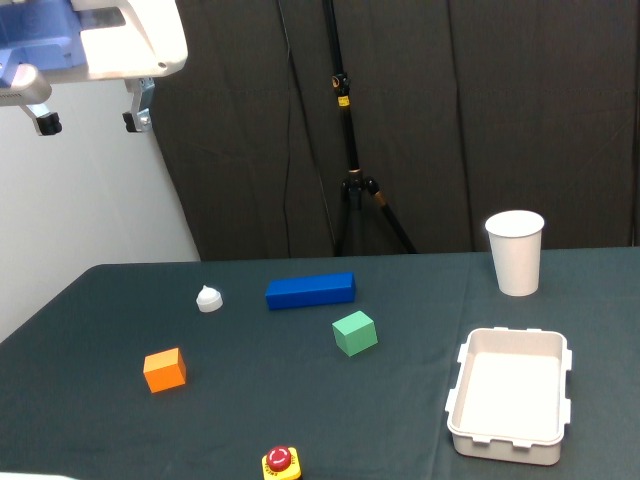}
        \caption{Right Third-person View}
        \label{fig:realworld_right}
    \end{subfigure}
    \hfill
    \begin{subfigure}[t]{0.32\textwidth}
        \centering
        \includegraphics[width=\linewidth]{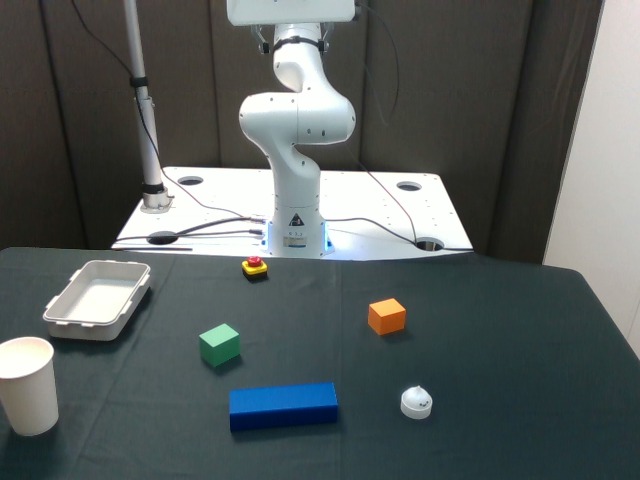}
        \caption{Front Third-person View}
        \label{fig:realworld_front}
    \end{subfigure}
    \caption{
    Camera views used in our real-world experiments.
    The right and front views provide third-person observations,
    while the wrist-mounted camera captures local interactions near the gripper.
    }
    \label{fig:realworld_views}
\end{figure*}

Our real-world setup uses three RGB cameras: a wrist-mounted
camera and two fixed cameras providing right and front
third-person views, as shown in Figure~\ref{fig:realworld_views}.
The wrist camera captures local visual details near the gripper,
while the third-person cameras provide complementary views
of the robot and workspace.
We combine the three views into a single visual input,
placing the wrist view above the two third-person views.
The same camera configuration and input layout are used
for the dense baseline and \algname{}.

\subsection{Action-to-Future Attention Entropy}
\label{app:action_entropy}
For action-layer selection, we measure the concentration of
action-to-future attention across all camera views and latent
timesteps. Unlike the view-dependent relevance scores used for
pruning, this calculation uses future-frame tokens for both
wrist and third-person views.

Following the notation in Section~\ref{sec:attention_pruning},
we first average the attention weights over heads and predicted
action tokens, then normalize over all $N_v=T_vH_vW_v$
future-video tokens:
\begin{equation}
p_t^{\mathrm{act},\ell}(i)
=
\frac{
\frac{1}{N_hN_a}
\sum_{h=1}^{N_h}
\sum_{j=1}^{N_a}
A_{a_j,f_{t,i}}^{(\ell,h)}
}{
\sum_{t'=1}^{T_v}
\sum_{u=1}^{H_vW_v}
\frac{1}{N_hN_a}
\sum_{h=1}^{N_h}
\sum_{j=1}^{N_a}
A_{a_j,f_{t',u}}^{(\ell,h)}
}.
\label{eq:appendix_action_future_probability}
\end{equation}
The normalized action attention entropy at layer $\ell$ is
\begin{equation}
\mathcal{H}^{\mathrm{act},\ell}
=
-\frac{1}{\log N_v}
\sum_{t=1}^{T_v}
\sum_{i=1}^{H_vW_v}
p_t^{\mathrm{act},\ell}(i)
\log p_t^{\mathrm{act},\ell}(i).
\label{eq:appendix_action_future_entropy}
\end{equation}
Here, $i$ spans spatial positions across all camera views,
and observation tokens are excluded from the normalization.
We use $0\log 0=0$, yielding
$\mathcal{H}^{\mathrm{act},\ell}\in[0,1]$.
Entropy is computed from the attention scores
during the initial conditional denoising forward pass.

\subsection{Entropy for phase identification}
\label{app:phase_id}
For phase identification, we use raw attention weights from
layer $\ell_{\mathrm{act}}$ during the initial conditional
denoising forward pass, without Gaussian smoothing.
Unlike the layer-selection entropy above, we compute entropy
separately for each action token and attention head before
averaging the resulting values.
In each case, attention weights are normalized over the
selected visual tokens, and entropy is divided by the
logarithm of the number of those tokens.

For $\mathcal{H}_{\mathrm{wrist}}$, we use the wrist-view
observation tokens and average the entropies over action
tokens and heads.
We compute observation entropy separately for each
third-person view in the same way and average across views
to obtain $\mathcal{H}_{\mathrm{third}}$.
For the temporal terms, each action token attends to the
full set of future-video tokens across all views and latent
timesteps.
After averaging these entropies over heads,
$\mathcal{H}_{\mathrm{early}}$ and
$\mathcal{H}_{\mathrm{late}}$ are obtained by averaging over
the first eight and last eight predicted action tokens,
respectively, in the 32-action chunk.

\subsection{Comparison of Token Selection Strategies}
\label{app:selection_comparison}
\begin{figure*}[!htbp]
\centering

\begin{subfigure}[t]{0.245\textwidth}
    \centering
    \includegraphics[width=\linewidth]
        {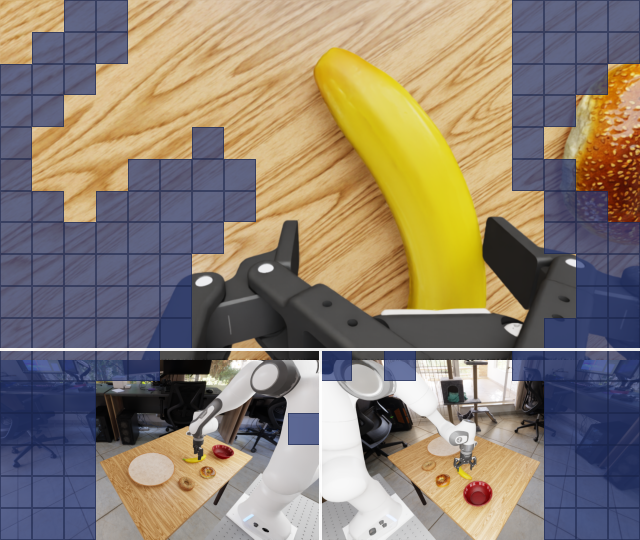}
    \caption{Instruction-only.}
    \label{fig:mask_instruction_only}
\end{subfigure}
\hfill
\begin{subfigure}[t]{0.245\textwidth}
    \centering
    \includegraphics[width=\linewidth]
        {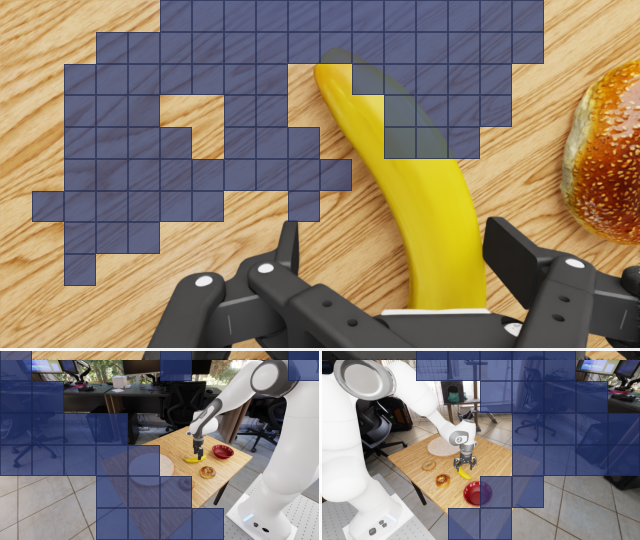}
    \caption{Action-only.}
    \label{fig:mask_action_only}
\end{subfigure}
\hfill
\begin{subfigure}[t]{0.245\textwidth}
    \centering
    \includegraphics[width=\linewidth]
        {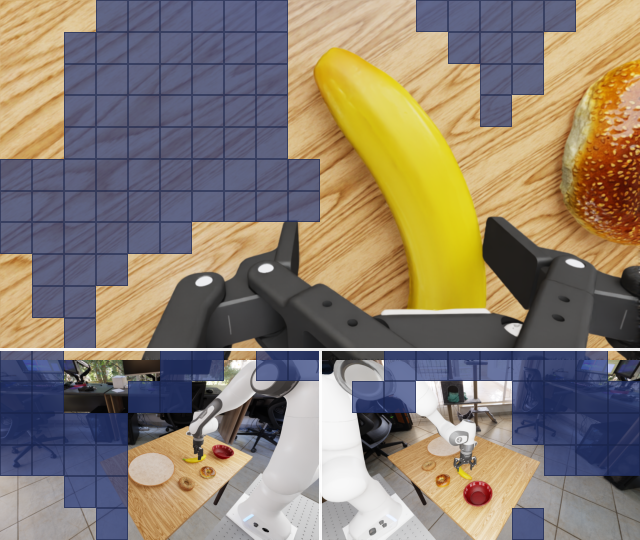}
    \caption{Instruction-first.}
    \label{fig:mask_instruction_first}
\end{subfigure}
\hfill
\begin{subfigure}[t]{0.245\textwidth}
    \centering
    \includegraphics[width=\linewidth]
        {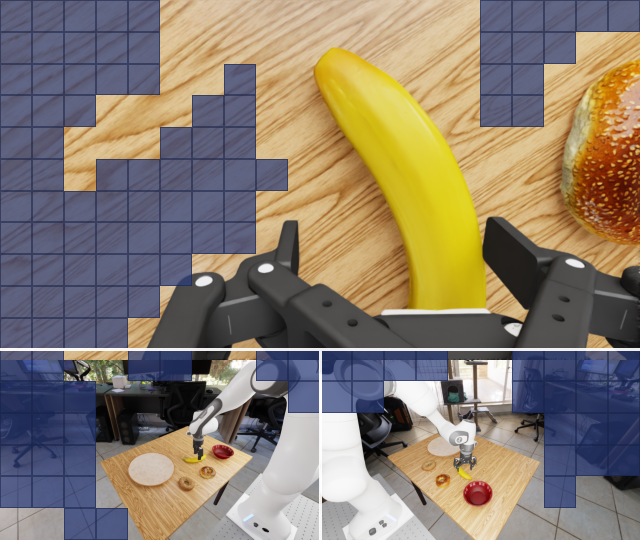}
    \caption{Action-first (Ours).}
    \label{fig:mask_action_first}
\end{subfigure}

\caption{
Comparison of the token selection strategies in Table~4
on the same RoboLab scene at 35\% pruning.
All variants use the same attention scores and
$5\times5$ Gaussian smoothing. Navy cells indicate pruned positions.
}
\label{fig:selection_mask_comparison}
\end{figure*}

Figure~\ref{fig:selection_mask_comparison} illustrates how
the selection strategy affects which regions are retained.
Action-only selection prunes part of the target object,
which may remove information needed to represent its shape
and position.
In contrast, instruction-only selection preserves the object
but removes part of the nearby gripper.
This may weaken the visual information describing the
interaction between the gripper and the object.
These examples illustrate why either attention source alone
may miss information relevant to manipulation.

Combining the two sources retains both object-related and
interaction-related regions, providing more complete coverage
of task-relevant visual information than either source alone.
Instruction-first and action-first selection produce broadly
similar masks, but instruction-first retains a tighter region
around the object in this example.
Action-first retains more surrounding context while preserving
parts of both the object and gripper.
This pattern suggests that prioritizing action relevance and
then adding instruction-relevant tokens can provide a useful
balance between interaction context and target-object information.
The quantitative comparison in Table~\ref{tab:ablation} further
supports the benefit of combining the two sources and evaluates
the effect of their selection order.

\subsection{Visualization of Pruned Tokens}
\label{app:pruning_visualization}

\begin{figure*}[t]
\centering
\begingroup
\setlength{\tabcolsep}{2pt}
\renewcommand{\arraystretch}{1.05}
\footnotesize

\newcommand{\pruninglabelwidth}{0.04\textwidth}

\newcommand{\pruningimagewidth}{%
  \dimexpr(\textwidth-\pruninglabelwidth-16pt)/4\relax}

\newcommand{\pruningratio}[1]{%
  \makebox[\pruninglabelwidth][c]{#1}}

\begin{tabular}{@{}ccccc@{}}
\toprule
\multicolumn{5}{c}{\textbf{RoboLab}} \\
\midrule
& \shortstack{Butter above\\raisin}
& \shortstack{Banana on\\plate}
& \shortstack{Canned food\\in bin}
& \shortstack{Stack\\bowls} \\
\addlinespace[3pt]
\pruningratio{35\%}
& \raisebox{-0.5\height}{\includegraphics[width=\pruningimagewidth]{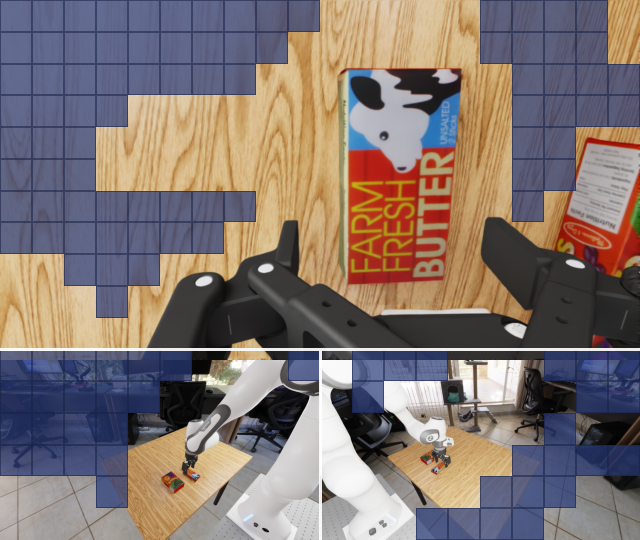}}
& \raisebox{-0.5\height}{\includegraphics[width=\pruningimagewidth]{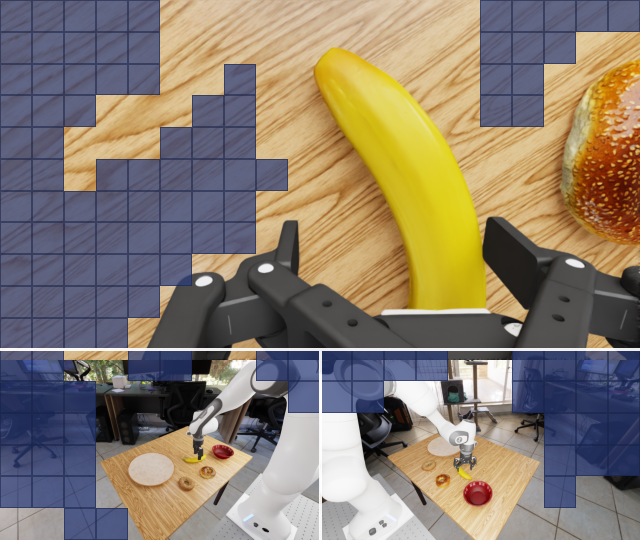}}
& \raisebox{-0.5\height}{\includegraphics[width=\pruningimagewidth]{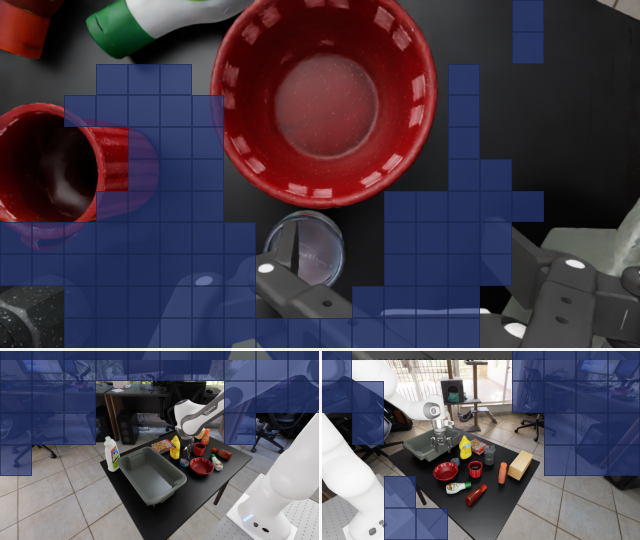}}
& \raisebox{-0.5\height}{\includegraphics[width=\pruningimagewidth]{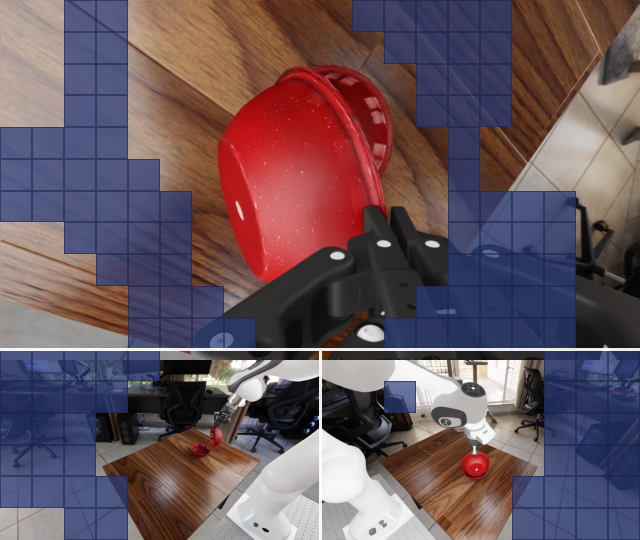}} \\
\addlinespace[8pt]
\pruningratio{50\%}
& \raisebox{-0.5\height}{\includegraphics[width=\pruningimagewidth]{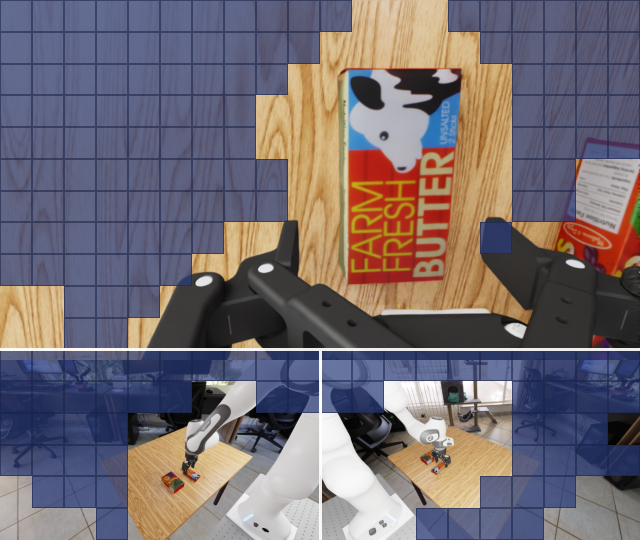}}
& \raisebox{-0.5\height}{\includegraphics[width=\pruningimagewidth]{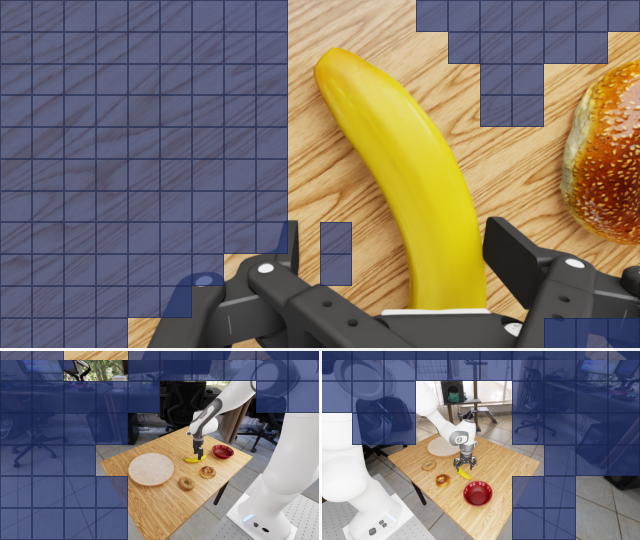}}
& \raisebox{-0.5\height}{\includegraphics[width=\pruningimagewidth]{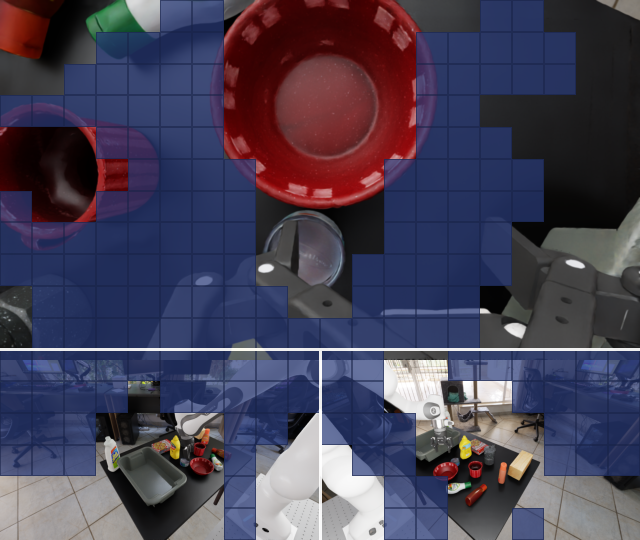}}
& \raisebox{-0.5\height}{\includegraphics[width=\pruningimagewidth]{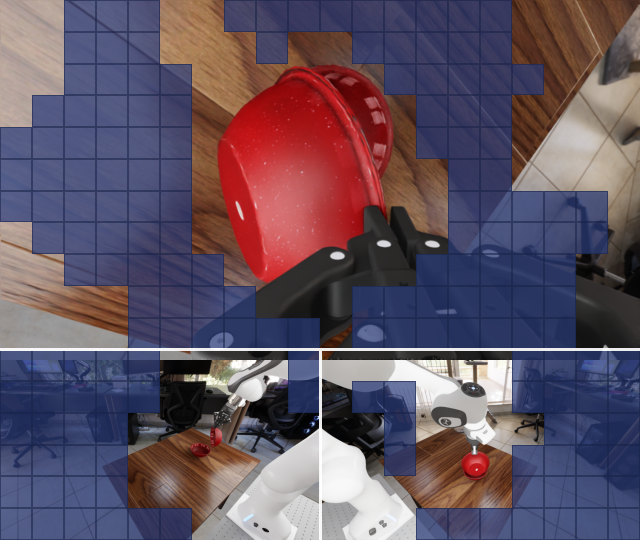}} \\
\bottomrule
\end{tabular}

\par\vspace{5pt}

\begin{tabular}{@{}ccccc@{}}
\toprule
\multicolumn{5}{c}{\textbf{RoboCasa}} \\
\midrule
& \shortstack{Open stand\\mixer head}
& \shortstack{Set up\\coffee mug}
& \shortstack{Turn on\\electric kettle}
& \shortstack{Navigate\\kitchen} \\
\addlinespace[3pt]
\pruningratio{35\%}
& \raisebox{-0.5\height}{\includegraphics[width=\pruningimagewidth]{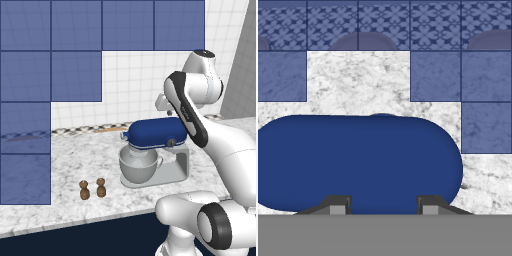}}
& \raisebox{-0.5\height}{\includegraphics[width=\pruningimagewidth]{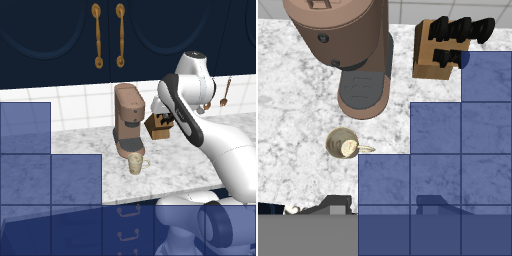}}
& \raisebox{-0.5\height}{\includegraphics[width=\pruningimagewidth]{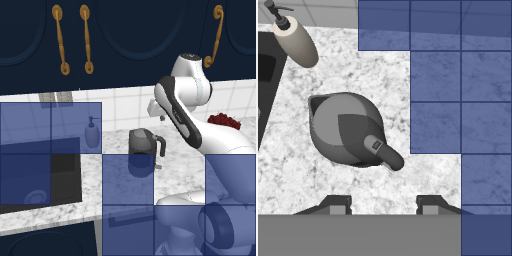}}
& \raisebox{-0.5\height}{\includegraphics[width=\pruningimagewidth]{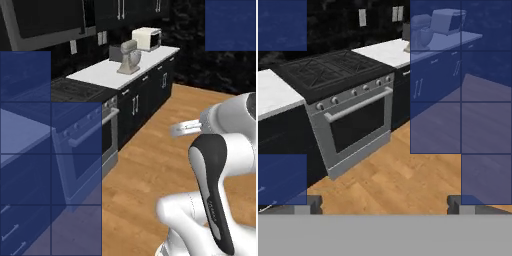}} \\
\addlinespace[8pt]
\pruningratio{50\%}
& \raisebox{-0.5\height}{\includegraphics[width=\pruningimagewidth]{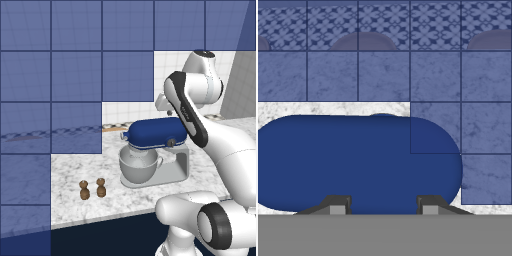}}
& \raisebox{-0.5\height}{\includegraphics[width=\pruningimagewidth]{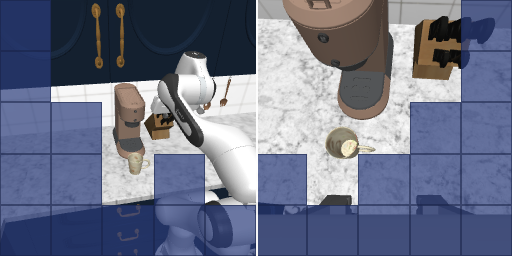}}
& \raisebox{-0.5\height}{\includegraphics[width=\pruningimagewidth]{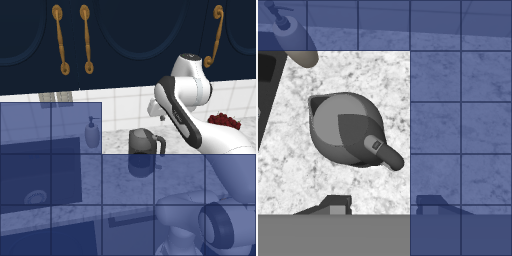}}
& \raisebox{-0.5\height}{\includegraphics[width=\pruningimagewidth]{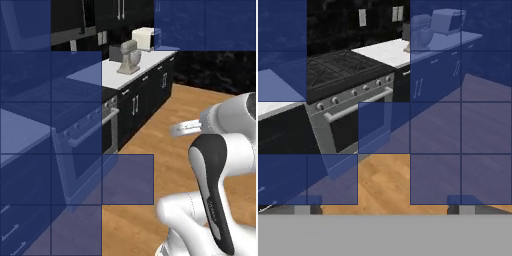}} \\
\bottomrule
\end{tabular}

\par\vspace{5pt}

\begin{tabular}{@{}cccc@{}}
\toprule
\multicolumn{4}{c}{\textbf{Real-world}} \\
\midrule
& \shortstack{Pick and\\place block}
& \shortstack{Stack\\cubes}
& \shortstack{Press\\button} \\
\addlinespace[3pt]
\pruningratio{35\%}
& \raisebox{-0.5\height}{\includegraphics[width=\pruningimagewidth]{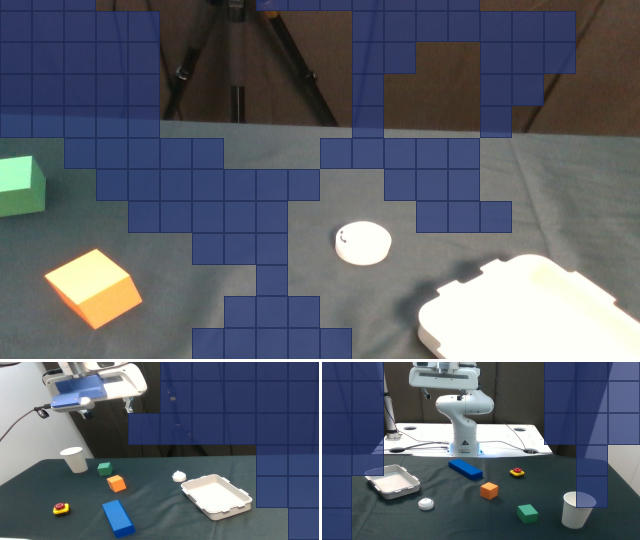}}
& \raisebox{-0.5\height}{\includegraphics[width=\pruningimagewidth]{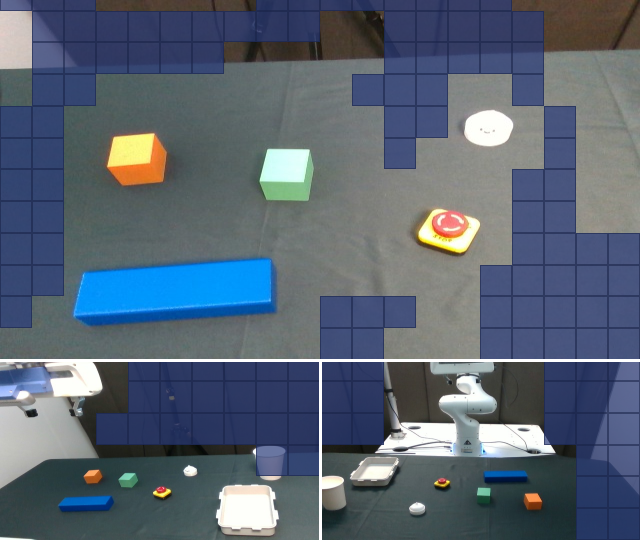}}
& \raisebox{-0.5\height}{\includegraphics[width=\pruningimagewidth]{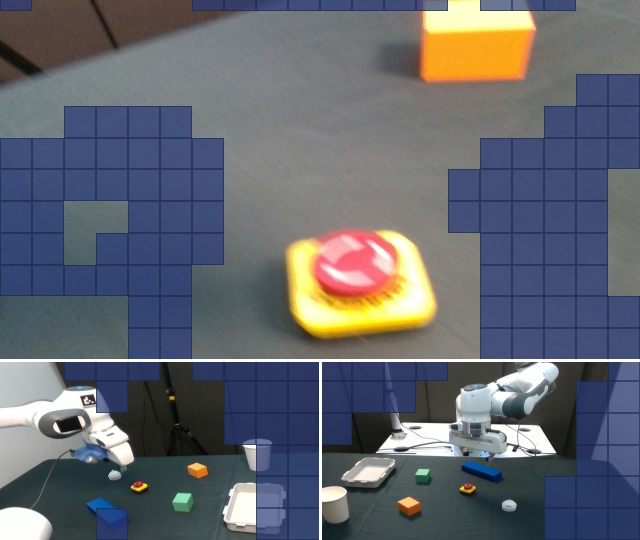}} \\
\addlinespace[8pt]
\pruningratio{50\%}
& \raisebox{-0.5\height}{\includegraphics[width=\pruningimagewidth]{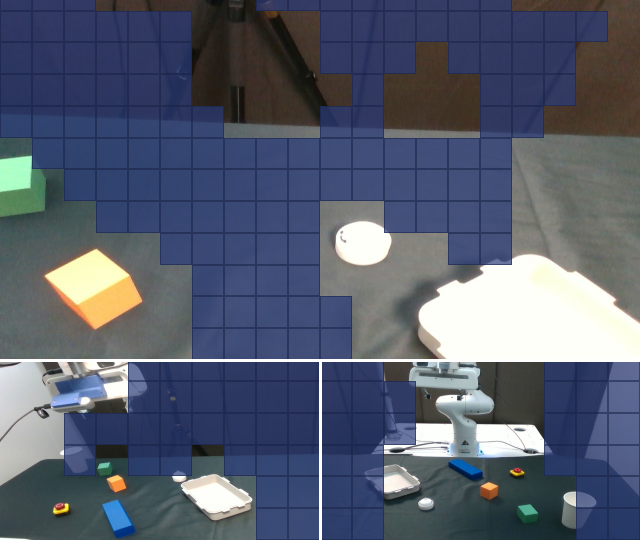}}
& \raisebox{-0.5\height}{\includegraphics[width=\pruningimagewidth]{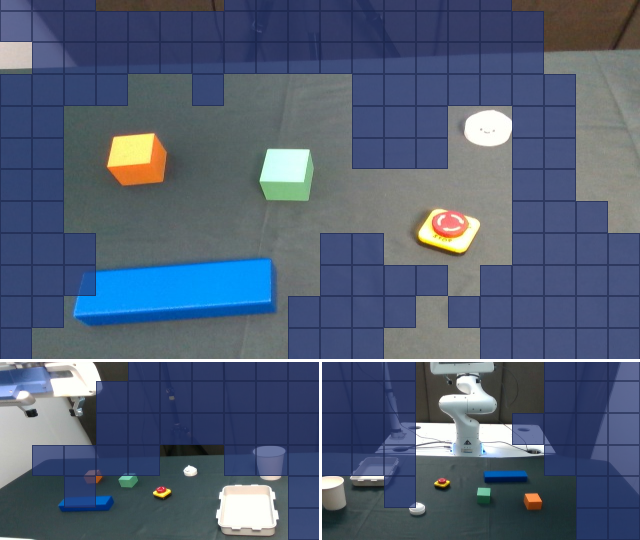}}
& \raisebox{-0.5\height}{\includegraphics[width=\pruningimagewidth]{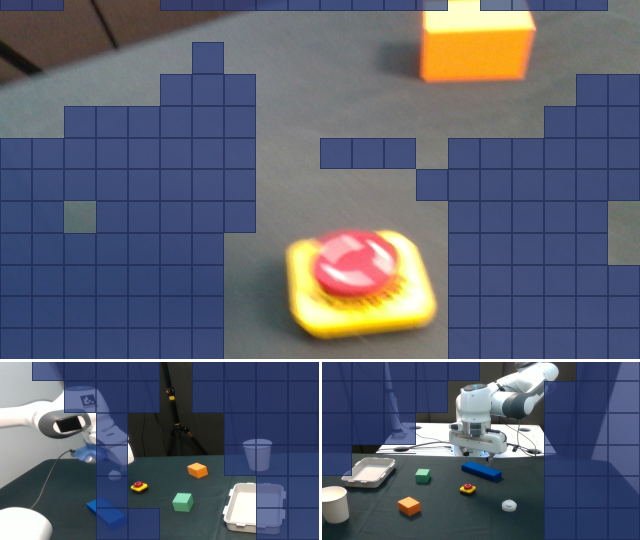}} \\
\bottomrule
\end{tabular}

\endgroup

\caption{
Examples of future-token pruning at 35\% and 50\% on RoboLab,
RoboCasa, and recorded real-world inputs. Navy cells indicate pruned positions.
}
\label{fig:appendix_pruning_examples}
\end{figure*}

Figure~\ref{fig:appendix_pruning_examples} compares
35\% and 50\% pruning on the same inputs from four RoboLab,
four RoboCasa, and three recorded real-world tasks.
The examples cover object manipulation, appliance interaction,
and navigation. In RoboLab, much of the butter package and banana remains
visible, while many surrounding regions are pruned.
The canned-food and bowl-stacking examples show similar
patterns around the objects and nearby workspace.
In RoboCasa, retained regions include parts of the robot,
appliances, and surrounding scene. The real-world examples also retain regions around the
blocks and emergency button at both pruning ratios.
Across these examples, the masks vary with the scene and
camera view rather than following a fixed spatial pattern.
At 50\%, more surrounding context is removed than at 35\%.

\clearpage
\subsection{Real-world rollouts}
\label{app:realworld_rollouts}
\begin{figure*}[!htbp]
    \centering
    \includegraphics[width=\textwidth]{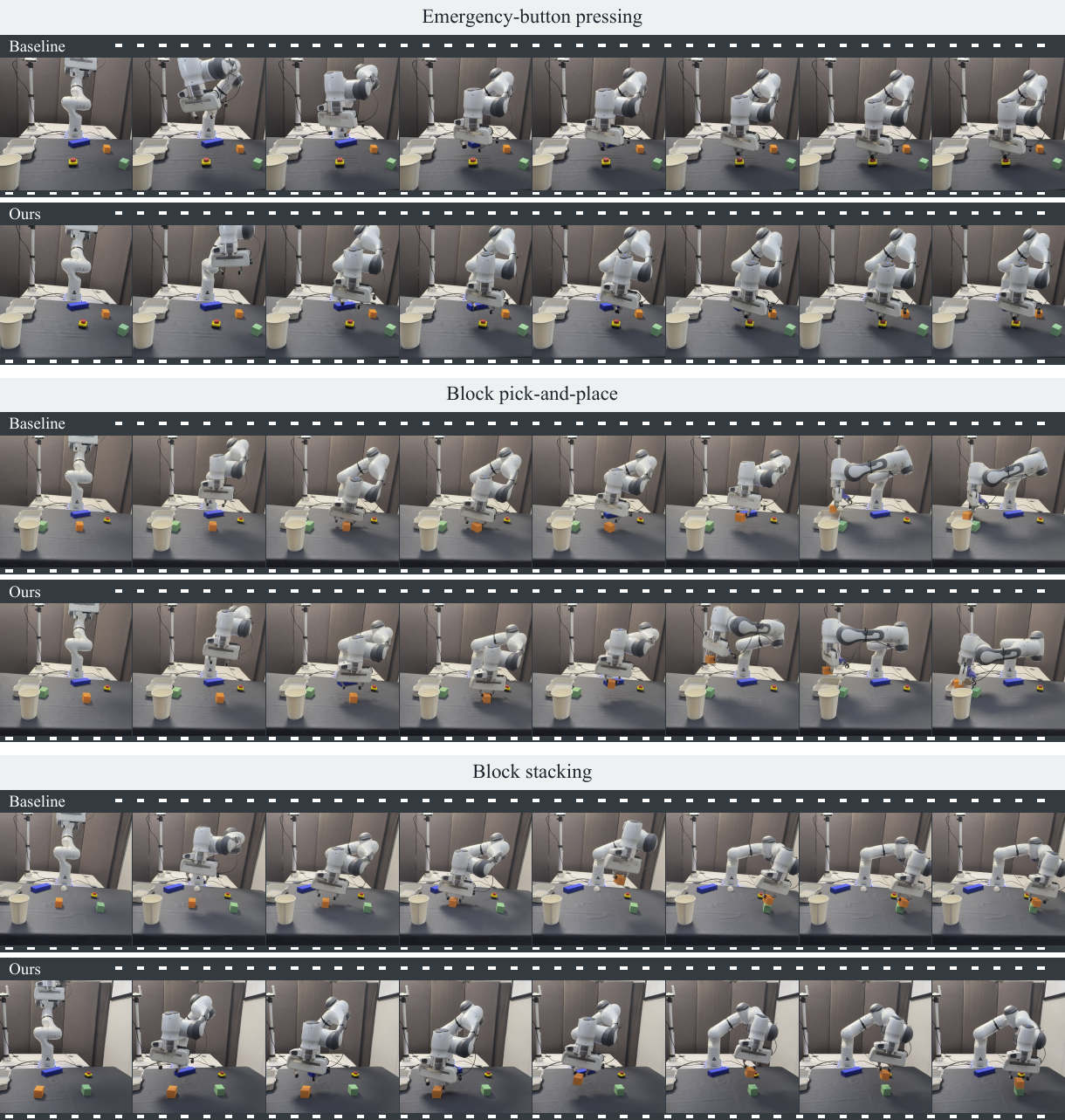}
    \caption{
        Representative real-world rollouts of the dense baseline
        and \algname{} on emergency-button pressing, block
        pick-and-place, and block stacking.
        Each row shows eight selected frames illustrating
        task progression from left to right.
    }
    \label{fig:realworld_rollouts}
\end{figure*}

Figure~\ref{fig:realworld_rollouts} shows representative
rollouts of the dense baseline and \algname{} on
emergency-button pressing, block pick-and-place, and
block stacking.
Each row contains eight selected frames showing how the
robot progresses through the task. For emergency-button pressing, the robot approaches the
button and brings the end effector into contact with it.
For block pick-and-place, the sequence shows the robot
reaching for the block, lifting it, and moving it toward
the bin.
Block stacking involves grasping a block, positioning it
above the other block, and lowering it into place.
These examples show both larger movements across the
workspace and local adjustments near the target.

The dense baseline and \algname{} follow similar overall
task progression, although their intermediate poses and
approach paths differ.
The examples illustrate how the pruned policy performs
the main stages of these tasks in a physical environment.
They complement the reported success rates by showing
the robot's behavior during execution.
Frames are selected independently for each rollout and
are not temporally aligned across methods; the figure
therefore does not provide a direct comparison of
execution time.

\end{document}